%% file: main-arxiv.tex
\documentclass{article}

\usepackage[nonatbib, preprint]{neurips_2026}

\input{math_commands.tex}

\usepackage[utf8]{inputenc} %
\usepackage[T1]{fontenc}    %
\usepackage[colorlinks=true, allcolors=blue]{hyperref}       %
\usepackage{url}            %
\usepackage{booktabs}       %
\usepackage{amsfonts}       %
\usepackage{nicefrac}       %
\usepackage{microtype}      %
\usepackage{xcolor}         %

\usepackage{svg}            %
\usepackage{amsmath}        %
\usepackage{xspace}
\usepackage{booktabs}
\usepackage{threeparttable}
\usepackage{siunitx}
\usepackage{booktabs}
\usepackage{threeparttable}
\usepackage{makecell}
 \usepackage[most]{tcolorbox}  %
\usepackage{multirow}
\usepackage{subcaption}     %
\usepackage{placeins}
\usepackage{mathtools}  %
\usepackage{colortbl}

\usepackage[style=numeric, natbib=true, url=false, sorting=none, maxbibnames=99]{biblatex}

\DeclareFieldFormat[article,inbook,incollection,inproceedings]{title}{#1}
\renewcommand{\bibfont}{\fontsize{9pt}{11pt}\selectfont}

\newcommand{\mname}{STROP\xspace}
\newcommand{\msmall}{\mname-S\xspace}
\newcommand{\mmedium}{\mname-M\xspace}

\title{Structure over Pixels: Learning Variable-Length Visual Programs}

\author{%
  Piotr Wyrwiński \\ %
  \textbf{Kacper Dobek} \\
  \textbf{Krzysztof Krawiec} \\
  Institute of Computing Science \\
  Poznan University of Technology, Poznan, Poland\\
  \texttt{piotr.wyrwinski@cs.put.poznan.pl} \\
}

\newcommand{\codefootnote}{}

\begin{document}

\maketitle

\input{abstract}

\input{body}

\input{ack}

\printbibliography

\newpage
\appendix

\input{appendix}

\end{document}

%% file: math_commands.tex
\usepackage{amsmath,amsfonts,bm}

\def\eqref#1{equation~\ref{#1}}

\def\ceil#1{\lceil #1 \rceil}

\def\1{\bm{1}}

\DeclareMathAlphabet{\mathsfit}{\encodingdefault}{\sfdefault}{m}{sl}
\SetMathAlphabet{\mathsfit}{bold}{\encodingdefault}{\sfdefault}{bx}{n}

%% file: abstract.tex
\begin{abstract}
Discrete visual tokenizers map images to ordered sequences of tokens, providing a natural representation for structural scene descriptions. Most use a fixed sequence length, while adaptive methods often require post-hoc search or choose among a small set of rates that control the length. We propose \mname, a discrete tokenizer that learns both a visual program and its image-dependent active length. A length head is trained with a four-phase curriculum using local rate--distortion probes against frozen DINOv3 features, then predicts the active prefix in a single forward pass.
At a matched rate of about $250$ nominal bits per crop, the adaptive model improves segmentation over a separately trained fixed-length baseline on four benchmarks (by $1.6$--$3.1$ mIoU), and it
also beats a fixed $K{=}32$ baseline that uses more bits.
\mname programs also yield higher segmentation mIoU than FlexTok, One-D-Piece, and ALIT at similar or higher rates, under the same readout architecture and training protocol.
\mname therefore learns useful per-image sequence lengths without post-hoc search or a predefined set of compression rates.
\end{abstract}

%% file: body.tex
\section{Introduction}
\begin{figure}[!ht]
    \centering
    \includegraphics[width=1\linewidth]{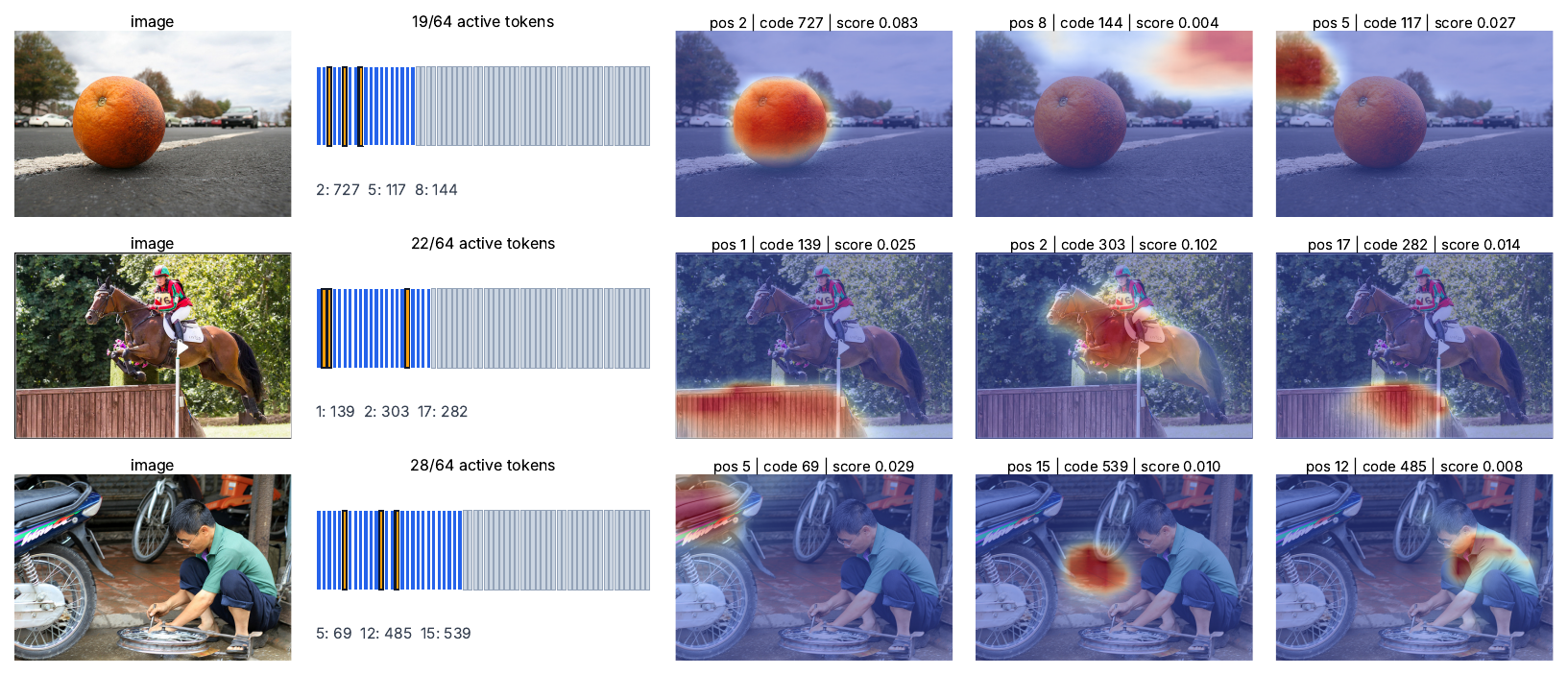}
      \caption{\textbf{Images as variable-length discrete 
      programs.}
  Each image is encoded as a 
  sequence of discrete 
  tokens
  whose length the model chooses per image.
  For each highlighted position, we erase the token and reinterpret the remaining sequence.
  The attribution panels show the resulting change in the DINO-aligned patch field.
  Red regions show that removing a token changes a spatially localized part of the patch field.
  Heatmaps are normalized per panel; scores are raw attribution magnitudes.}
    \label{fig:teaser}
\end{figure}

Deep neural networks prove unmatched at most computer vision (CV) tasks, mainly thanks to efficient representation learning. Their learned representations remain however verbose, distributed, convoluted, and thus hard to inspect and validate in use cases that require robust scene interpretation, like robotics, planning, autonomous driving or medical diagnosis, which require models that reason about scenes in terms of objects, their properties, and their spatial relations. Yet such structural visual descriptions remain hard to learn, especially when supervision is scarce. 
Object-centric learning~\citep{locatello2020slot,singh2022slate,seitzer2023dinosaur} and neural program induction~\citep{reed2016npi} both pursue such descriptions.

\emph{Discrete visual tokenizers} (DVTs) \citep{dvt-survey2026} offer a promising framework for forming structural descriptions, since they already operate with quasi-symbols: a scene is mapped to an ordered sequence of discrete tokens from a fixed vocabulary, which can be usually interpreted back into the spatial feature field. The model learns to generate such sequences without object labels, hand-designed predicates, or a-priori helper structures like scene graphs. Enforcing such latent representations facilitates inspection in search of signs of code reuse, symbol grounding, or compositionality.

Most DVTs operate at constant sequence length. VQ-VAE and VQGAN-style models produce a fixed spatial grid of tokens, and recent 1D tokenizers replace that grid with a fixed number of latent slots~\citep{vandenoord2017vqvae,esser2021vqgan,yu2024titok}. A smooth sky patch and a crowded street are mapped to different sequences, but both receive the same length (sometimes referred to as token \emph{budget}). This sits uneasily with cognitive postulates like Gestalt principles and Occam's razor, not to mention the formalisms of minimum description length and Kolmogorov complexity (KC), which ties the complexity of an artifact to the length of the shortest program that generates it~\citep{kolmogorov1965,li2008kolmogorov}. Uncomputable in general, KC nonetheless makes the right demand: adding an object to a scene should come at a cost. The sequence length should correlate with the `amount' of visual structure in the observed scene -- this is the first step, if not a prerequisite, for forming principled structural representations.

In this study, we call a sequence of tokens a \emph{program}
in a deliberately thin sense. The codebook acts as a fixed instruction set, but the interpreter learns what each instruction does, and no grammar restricts how instructions combine. Neural program induction, by contrast, usually starts from predefined operations~\citep{reed2016npi}.
\emph{Incremental interpretability} is our only requirement, i.e. every prefix of a program is a valid input to the same interpreter. %
Some researchers train decoders to tolerate prefixes; others use inference-time search, recurrent allocation, routing among fixed rates, or information-theoretic scores of input compressibility~\citep{bachmann2025flextok,miwa2025onedpiece,yan2025elastictok,duggal2025alit,shen2025cat,ye2025infotok}. Closest to us is KARL~\citep{duggal2025karl}, which predicts a per-image length in one forward pass and reads it as an estimate of complexity, though conditioned on a pixel-space reconstruction quality supplied at inference rather than on the tokenizer's own economy. These results suggest that length control should be tightly linked with visual tokenization, rather than being delegated to some external mechanism. %

Another prevailing characteristic in DVT research is the emphasis on fidelity of reconstruction. Indeed, previous works suggest that training DVTs with pixel-level reconstruction loss serves compression well. However, it can waste budget on capturing textural characteristics that carry little semantic or geometric structure. Feature-distillation tokenizers like BEiT~v2 \citep{peng2022beitv2}, DINOSAUR \citep{seitzer2023dinosaur}, REPA \citep{yu2025repa}, and MAETok \citep{chen2025maetok} show that guiding training with teacher features and representation-aligned objectives, rather than just reconstruction loss, can produce useful latent spaces and object-sensitive structure.

Motivated by the above findings, we propose \mname\codefootnote, a DVT architecture 
built from two transformers. A generator writes a sequence of tokens from a visual vocabulary, and an interpreter reads it back into a patch feature field. A learned length head decides how much of the sequence to expose. Each token costs a fixed number of bits, so this length sets the \emph{rate} of the encoding, which the model trades against the distortion of the interpreted features.
This gives three concrete contributions:
\textbf{(i) Variable-length visual programs.} We supervise per-image program length with local rate-distortion probes of the trained tokenizer and use a prefix curriculum -- random truncation first, predicted prefixes later.
\textbf{(ii) A simple generator-interpreter tokenizer.} Two encoder-only transformers with learned queries write and read causally ordered visual tokens. DINOv3 features provide the only supervision.
\textbf{(iii) Evaluation of representation and structure.}
We compare adaptive programs with fixed-length tokenizers and public adaptive tokenizers, ablate the training design, measure downstream dense prediction and analyze where tokens act in the image and how program length follows scene content.

\section{The proposed \mname architecture}

\input{body/figure-architecture}

The architecture is shown in Figure~\ref{fig:arch-diagram}. Given an input image $x \in \mathbb{R}^{3 \times H \times H}$, we seek a
compact discrete program $Z_q = (z_{q_{1}}, \dots, z_{q_{L}})$
from which a DINO-aligned feature field can be recovered,
where $L$ varies by image and
$\mathcal{V}=\{v_i\in\mathbb{R}^{d_c}\}_{i=1}^{V}$
is a visual vocabulary of size $V$.
The model comprises a frozen \emph{visual
encoder} producing patch-wise features, a \emph{program generator} $G$ that maps those features into a
sequence of tokens, a vector-quantization bottleneck, and an
\emph{interpreter} $I$ that expands the program back to a spatial feature grid. An auxiliary \emph{length head} predicts $L$ from
the encoder features, enabling variable-length programs at inference
without autoregressive sampling.

\input{body/architecture-components}

\subsection{Training objective}
\label{sec:training-objective}

\input{body/training-objective}

\subsection{Adaptive program-length curriculum}
\label{sec:curriculum}

\begin{figure}
    \centering
    \includegraphics[width=\linewidth]{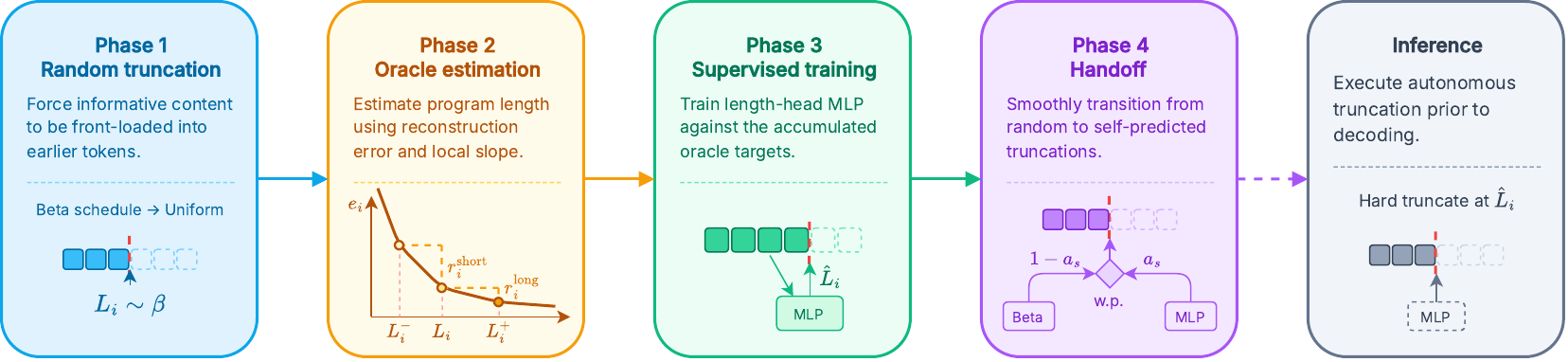}
    \caption{Training curriculum for adaptive truncation. \textbf{Phase~1:} Random Beta truncation encourages front-loading of information. \textbf{Phase~2:} Oracle lengths are estimated from the feature reconstruction error and the local slope of the rate--distortion curve. \textbf{Phase~3:} An MLP length head is supervised on these oracle targets. \textbf{Phase~4:} A probabilistic handoff smoothly transitions from random to MLP-predicted lengths. \textbf{Inference:} Hard truncation relies solely on the learned MLP. Details on the curriculum can be found in Appendix \ref{app:curriculum-details}.  
}
    \label{fig:curriculum}
    \vspace{-3mm}
\end{figure}

The generator produces \(K\) quantized tokens for each sample \(x\). The interpreter uses only a prefix of length \(L_i \in \{1,\dots,K\}\) and masks positions \(k>L_i\). We train the length head with the four-phase curriculum shown in Fig.~\ref{fig:curriculum}. Appendix~\ref{app:curriculum-details} provides more details on the schedule.\\
\noindent\textbf{Phase 1: Random truncation.}
We train the model on randomly truncated programs. The length distribution favors longer prefixes, and the generator learns to place useful information in the early tokens. The length head remains inactive.\\
\noindent\textbf{Phase 2: Oracle target estimation.}
We continue random truncation and estimate a target length for each sample
from its feature reconstruction error and from the local rate--distortion slope, using slightly shorter and longer prefixes.
These targets are collected but do not yet affect truncation.\\
\noindent\textbf{Phase 3: Supervised length-head training.}
We train the length head on the targets estimated in Phase 2. Prefix lengths are still sampled independently of its predictions.\\
\noindent\textbf{Phase 4: Handoff to predicted lengths.}
We gradually replace the sampled lengths with the predictions from the length head. After the handoff, all programs use predicted lengths, while the head continues to receive supervision. Appendix~\ref{app:stability} analyzes sensitivity to the handoff timing.
At inference, we round \(\hat{L}_i\), clip it to \([1,K]\), and pass only the resulting prefix to the interpreter.

\input{body/figure-unsupervised-redouts-coco}

\input{body/related-work}

\begin{table}[t]
\centering
\caption{Rate-matched comparison with fixed-length tokenizers. \textbf{Segmentation mIoU} (\%, $\uparrow$) on four benchmarks. All rows use a frozen DINO-S encoder, the \msmall generator/interpreter and COCO training; each fixed-length baseline is trained separately with $K$ tokens and no length head. Bits are nominal index bits per image: 10 bits per token, plus a 6-bit length field for \mname. The $K{=}25$ baseline is rate-matched to \mname via $K{=}\lfloor \bar{L} + 6/\log_2 V + 0.5 \rfloor$, where $\bar{L}$ is the mean predicted program length on COCO val. mIoU values are means over 3 tokenizer seeds, each averaged over 3 probe seeds; probes are trained on the full training split at $256{\times}256$ for 40k updates. Latent probes observe the interpreter's spatial latent field with a $1{\times}1$ head; program probes process the active program token vectors $z_q$ with a lightweight transformer (3.8M parameters). 
DINO-S probes the frozen teacher features that \mname learns to reproduce, as a reference without compression.
Best values among the ${\approx}250$-bit tokenizers in bold.}
\label{tab:T1}
\setlength{\tabcolsep}{4pt}
\renewcommand{\arraystretch}{1.1}
\begin{tabular*}{\linewidth}{@{\extracolsep{\fill}}lrlrrrr@{}}
\toprule
Model & Bits & Representation & VOC & COCO-Stuff-27 & ADE20K & Citysc. \\
\midrule
Teacher (frozen DINO-S) & --- & teacher & 80.3 & 46.2 & 42.1 & 56.7 \\
\midrule
Fixed-length $K{=}25$ & 250 & latent  & 66.3 & 39.4 & 25.0 & 33.9 \\
               &     & program & 59.5 & 37.7 & 23.4 & 35.7 \\
\cmidrule(l){1-7}
Fixed-length $K{=}32$ & 320 & latent  & 67.0 & 39.5 & 25.8 & 34.9 \\
               &     & program & 60.1 & 37.9 & 23.7 & 35.9 \\
\cmidrule(l){1-7}
Fixed-length $K{=}64$ & 640 & latent  & 73.0 & 43.4 & 32.3 & 42.0 \\
               &     & program & 67.6 & 42.2 & 29.5 & 42.9 \\
\midrule
\mname (ours, adaptive) & ${\approx}$250 & latent  & \textbf{68.6} & \textbf{41.2} & \textbf{28.0} & 36.8 \\
                  &                & program & 61.5 & 39.3 & 25.3 & \textbf{37.8} \\
\bottomrule
\end{tabular*}
\end{table}

\section{Results}
\input{body/results}

\input{body/limitations}

\section{Conclusion}

We introduced \mname, a discrete visual tokenizer that predicts an
image-dependent program length using local rate--distortion supervision
against frozen DINOv3 features.
At a matched rate, its adaptive programs keep
more segmentation-relevant information than separately trained fixed-length
programs, and with a shared readout they outperform public adaptive
tokenizers. The gain depends mostly on training against features without
pixel gradients and on the staged length curriculum, and it holds with
ImageNet training and with a DINO-B teacher. Program length follows scene
complexity, and erasing a single token changes a localized part of the
interpreted field.

%% file: body/figure-architecture.tex
\begin{figure}[t]
    \centering
    \includegraphics[trim=2pt 0pt 2pt 10pt, clip, width=\linewidth]{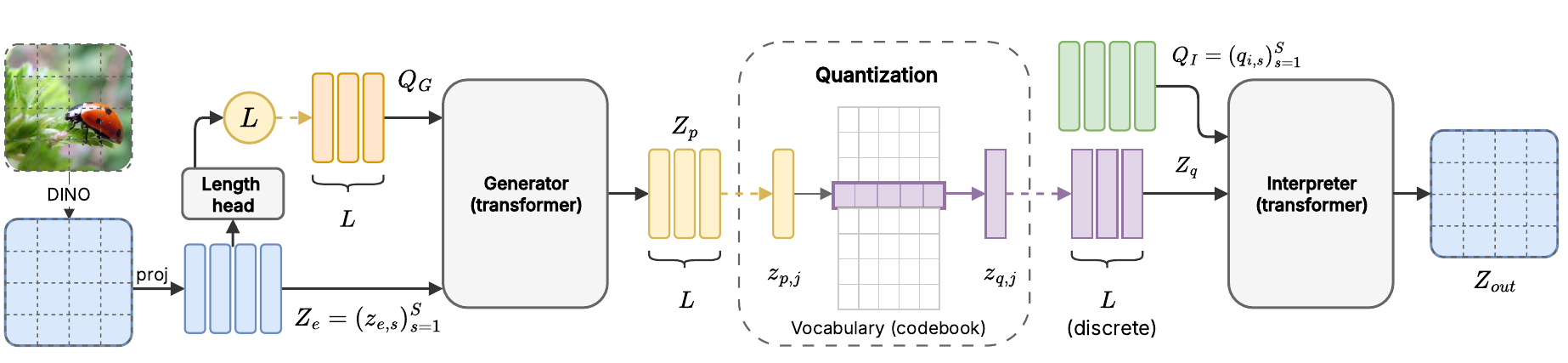}
    \caption{\mname architecture overview.}
    \label{fig:arch-diagram}
    \vspace*{-4.5mm}
\end{figure}

%% file: body/architecture-components.tex
\textbf{Encoder.}
We extract token-level features with a frozen DINOv3 ViT
encoder~\citep{simeoni2025dinov3}. For patch size $p$, it produces $S = (H/p)^2$ tokens. A layer norm followed by a linear projection
maps every token from $d_{\text{enc}}$ to a shared model dimension $d$,
yielding source features $Z_e \in \mathbb{R}^{S \times d}$. %

\textbf{Program generator.}
To map the patch-based visual features into a sequence of tokens describing the scene, we adopt an encoder-only
transformer that processes $Z_e$ with an appended sequence $Q_G$ of $K$ learnable query tokens. The concatenated $[Z_e;\, Q_G] \in \mathbb{R}^{(S+K) \times d}$
is processed by standard transformer encoder layers. A causal mask
prevents the $k$th query token from attending to query tokens with $k' > k$, and
the source tokens from attending to any query, ensuring that no future-query
information leaks through intermediate source representations. After
encoding, only the $K$ query outputs are retained as the pre-quantized, raw
\emph{program} $Z_p \in \mathbb{R}^{K \times d}$.

\textbf{Vector quantization.}
A linear projection maps each token of $Z_p$ into the codebook space
$\mathbb{R}^{K \times d_c}$ ($d_c \ll d)$, where an $V$-entry vocabulary assigns
discrete indices via nearest-neighbor lookup using Euclidean distance. The quantized tokens are
projected back to $d$ dimensions, yielding $Z_q \in \mathbb{R}^{K \times
d}$. %

\textbf{Length head.}
A lightweight MLP $h_\phi$ predicts the program length directly from the
encoder features: $\hat{L} = \ceil{K\sigma\!\bigl(h_\phi(\mathrm{pool}
(Z_e))\bigr)}$, where $\mathrm{pool}$ denotes mean pooling over the
spatial dimension and $\sigma$ is the sigmoid function. At inference,
$\hat{L}$ is rounded to the nearest integer. Positions $k > \hat{L}$
in $Z_q$ are masked from the interpreter. Training of this
head is detailed in Section~\ref{sec:curriculum}.

\textbf{Interpreter.} The interpreter mirrors the generator in mapping $Z_q$ truncated to $\hat{L}$ tokens back to patch-level feature tokens. Padding is used only in training, where selected prefixes are padded for batching and masked so that the interpreter cannot attend to padded positions; no padded or zeroed program positions are passed at inference.  The interpreter updates $(H/p)^2$ learnable grid queries by cross-attending to the retained tokens, treating the selected program as the key-value memory. The resulting grid-query outputs are projected with a $1{\times}1$ convolution and reshaped into a feature map ${Z_{out}} \in \mathbb{R}^{(H/p) \times (H/p) \times d}$ to obtain the patch-feature field.

\textbf{Relation to feature autoencoders.}
\mname reconstructs frozen DINO features, so it can be viewed as a discrete feature autoencoder. A standard feature autoencoder uses a fixed-size latent representation. \mname instead produces a causally ordered sequence and predicts how much of it to retain. Any prefix is a valid input to the interpreter, and the predicted length controls how much information passes through the bottleneck. The variable-length bottleneck and the vocabulary are learned together.

%% file: body/training-objective.tex
The loss function is a composite of the following objectives. \\
\textbf{Latent alignment} $\mathcal{L}_{\mathrm{lat}}$.
Given the interpreter-generated patches $Z_{out}$ and the frozen DINO patches $Z_e$, we use
$\mathcal{L}_{\mathrm{lat}} = 1-\cos(Z_{out}, Z_e) +\mathrm{MSE}(Z_{out}, Z_e)$,
where the cosine term preserves the teacher-feature geometry and MSE stabilizes scale.\\
\textbf{Commitment loss} $\mathcal{L}_{\text{commit}}$. The standard VQ regularization penalty $\mathcal{L}_{\text{commit}}=\lambda_{\text{q}}\|Z_p - \mathrm{sg}[Z_q]\|^2$  keeps pre-quantized embeddings close to their selected codebook entries.\\
\textbf{Diversity loss} $\mathcal{L}_{\text{div}}$. The variance and covariance terms of VICReg~\citep{bardes2022vicreg}, applied to the pre-quantization program vectors $Z_p$: a hinge keeps each embedding dimension's variance from collapsing, and an off-diagonal covariance penalty decorrelates dimensions ($c{=}0.04$), together giving the codebook distinct directions to specialize into. We omit VICReg's invariance term, which requires paired views we do not have. Warmed up over $40$k steps with weight $\lambda_{\mathrm{div}}{=}0.3$.\\
The full training objective is thus:
\(
\mathcal{L} = \mathcal{L}_{\text{lat}} + \mathcal{L}_{\text{commit}} + \mathcal{L}_{\text{div}} + %
\mathcal{L}_{\text{len}},
\)
where $\mathcal{L}_{\text{len}}$ (Sec.~\ref{sec:curriculum}) is activated in Phase~3 of the curriculum. The weights for each term are given in Table~\ref{tab:training}.\\ %
\textbf{Codebook design.}
The vocabulary consists of $V{=}1024$ entries of $d_c{=}16$ dimensions, updated via exponential moving average (EMA, $\gamma{=}0.95$) with $\ell_2$-normalized codes. The low code dimension combined with $\ell_2$ normalization places all entries on a hypersphere, stabilizing nearest-neighbor assignment and preventing dead codes from drifting far away from the active manifold.

%% file: body/figure-unsupervised-redouts-coco.tex
\begin{figure}
    \centering
    \includegraphics[width=1.0\linewidth]{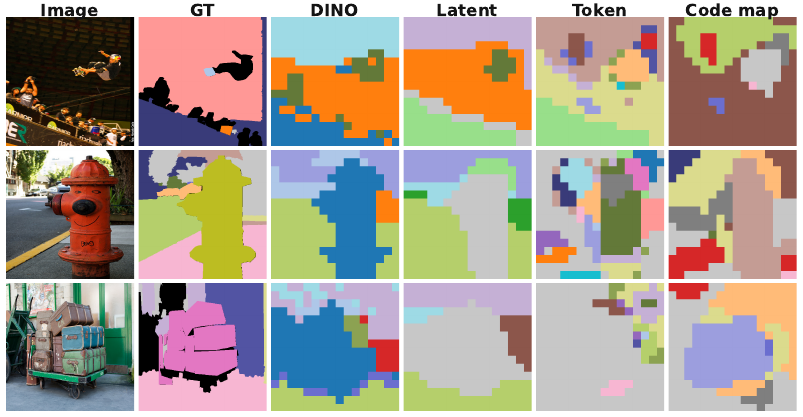}
    \caption{Unsupervised readouts on COCO-Stuff-27. DINO, Latent, and Token show k-means regions obtained from DINO teacher patch features, \mname latent patch features, and token-erasure attribution vectors, respectively. Colors denote arbitrary cluster IDs and are
  not semantic labels. Code map shows a many-to-one mapping from program code IDs to semantic classes.
}
    \label{fig:unsupervised-segmentation}
\end{figure}

%% file: body/related-work.tex
\section{Related work}
\label{sec:related}

\textbf{Discrete visual tokenizers.}
VQ-VAE introduced discrete visual representations based on a continuous encoder and learned codebook~\citep{vandenoord2017vqvae,razavi2019vqvae2}, while VQGAN added adversarial and perceptual losses~\citep{esser2021vqgan}. Later work developed masked and next-scale prediction~\citep{chang2022maskgit,tian2024var}, alternative quantizers~\citep{mentzer2024fsq,lee2022rqvae}, and a more stable straight-through estimator~\citep{fifty2024rotation}. LFQ in MAGVIT-v2 replaces the learned codebook with binarized latent dimensions~\citep{yu2024magvitv2}; BSQ first normalizes the latents onto a hypersphere~\citep{zhao2025bsq}. Continuous and product-quantized alternatives also shorten autoregressive sequences~\citep{chen2024softvq,li2024imagefolder}. These methods retain a fixed token count determined by the spatial grid.

\textbf{Variable-length tokenization and adaptive computation.}
TiTok showed that images can be represented by a short sequence of 1D tokens~\citep{yu2024titok}. FlexTok trains a rectified-flow decoder with nested dropout~\citep{bachmann2025flextok}, and One-D-Piece applies tail-token dropout to a discrete tokenizer~\citep{miwa2025onedpiece}. ALIT adds latents through recurrent rollouts~\citep{duggal2025alit}; ElasticTok selects the length using a reconstruction-quality threshold~\citep{yan2025elastictok}; CAT routes images among three compression ratios~\citep{shen2025cat}; and InfoTok uses an ELBO-based router~\citep{ye2025infotok}.

KARL predicts per-token halting in one pass, conditioned on a reconstruction-loss target supplied at inference~\citep{duggal2025karl}. \mname also predicts length in one pass. Its supervision comes from the local rate--distortion slope measured against frozen teacher features, so inference does not require an external quality target. KARL uses token sets, while \mname a causally ordered prefix.

\textbf{Reducing tokens inside a fixed encoder.}
DynamicViT prunes input tokens through predictors inserted at several transformer layers~\citep{rao2021dynamicvit}, while ToMe merges similar tokens without additional training~\citep{bolya2023tome}. Foveated inference instead reduces the input resolution away from fixation~\citep{harari2025foveation}. These methods remove, merge, or subsample input tokens. \mname produces a sequence of learned codebook entries that is not a subset of the input.

\textbf{Feature-space reconstruction.}
I-JEPA predicts continuous, fixed-size spatial representations rather than pixels~\citep{assran2023ijepa}. BEiT~v2 distills a CLIP teacher into discrete codes~\citep{peng2022beitv2}, while RCG learns a diffusion prior over self-supervised representations~\citep{li2024rcg}. REPA aligns diffusion-transformer states with frozen DINOv2 features~\citep{yu2025repa}, and REPA-E extends this signal through the VAE tokenizer~\citep{leng2025repae}. Related tokenizers regularize VAE latents toward vision-foundation-model features~\citep{yao2025vavae}, use masked feature prediction~\citep{chen2025maetok}, or formulate training as latent denoising~\citep{yang2025ldetok}. \mname instead distills frozen DINOv3 features into a discrete sequence whose length varies by image.

\textbf{Compositional and object-centric representations.}
Slot Attention uses iterative competitive attention to form object-centric representations~\citep{locatello2020slot}. SLATE decodes slots autoregressively over discrete tokens~\citep{singh2022slate}, and DINOSAUR reconstructs self-supervised features to obtain slots on real images~\citep{seitzer2023dinosaur}. SlotDiffusion combines slots with latent diffusion~\citep{wu2023slotdiffusion}, while SlotFormer extends object-centric representations to video~\citep{wu2023slotformer}. Wen et al.~\citet{wen2025principal} impose a PCA-like order on 1D tokens, and NEO learns latent programs executed by a shared transition model~\citep{baek2026neo}. \mname uses a causally ordered prefix rather than a permutation-invariant set, which makes program length meaningful.

\textbf{Generator--interpreter framing.}
Our terminology follows Neural Programmer-Interpreters~\citep{reed2016npi}. The architecture uses learned queries to compress and reconstruct spatial representations, as in DETR~\citep{carion2020detr}, Perceiver~\citep{jaegle2021perceiver}, and Perceiver IO~\citep{jaegle2022perceiverio}. \mname adds a variable-length prefix with end-to-end length supervision.

%% file: body/results.tex
\subsection{Experimental setup} 
We train five model variants by sweeping the visual encoder's (DINOv3 ViT-S/B/L~\citep{simeoni2025dinov3}) and generator/interpreter's capacity (small \msmall, medium \mmedium). The encoder is always frozen; trainable components comprise the generator, interpreter, quantizer projections, and length head.
Unless otherwise stated, the generators produce programs of length $K{=}64$ with 2D patch and learned prefix positional encodings, a vector quantizer with $V{=}1024$ codes of dimension $d_c{=}16$, $\ell_2$-normalized code vectors, EMA updates with $\gamma{=}0.95$, and commitment weight $\lambda_{\text{q}}{=}1.0$.
\msmall uses $d_{\mathrm{model}}=256$, 8 layers, and an FFN width of 1024, while \mmedium uses $d_{\mathrm{model}}=512$, 10 layers, and an FFN width of 2048.
The interpreter mirrors the generator in every variant, sharing $d_{\text{model}}$, depth, number of heads (always 8), and feed-forward width. Table~\ref{tab:variants} in Appendix~\ref{app:variants} lists all configurations with parameter budgets and training cost.
Unless otherwise stated, training engages curriculum learning of the length head (Sec.\ \ref{sec:curriculum}, Appendix \ref{app:curriculum-details}), parametrized according to Table~\ref{tab:curriculum}: Phase 1 lasts 100k and 200k steps for \msmall and \mmedium respectively, and each of the subsequent three phases lasts 50k steps.\\
\textbf{Datasets.}
All five variants are trained on both ImageNet-1k~\citep{imagenet2009} and COCO 2017~\citep{coco2014} at $256{\times}256$ resolution. We additionally train on CLEVR~\citep{clevr2017} at $128{\times}128$ with reduced program length ($K{=}32$) and smaller codebooks ($V \in \{32, 64, 128\}$) to study codebook sizing and program structure on a controlled scene domain (Appendix~\ref{app:clevr}).\\
\textbf{Optimization.}
All variants share the same optimizer and schedule: AdamW with peak learning rate $10^{-3}$ (length head: $3{\times}10^{-4}$), batch size 32, and $5{\times}10^{5}$ total steps. The learning rate follows a warmup-hold-cosine schedule (30k warmup, hold to 100k, cosine decay to $10^{-4}$). The rate-matched comparison and the ablations (Tables~\ref{tab:T1} and~\ref{tab:T3}) use the COCO-trained DINO-S/\msmall configuration with three tokenizer seeds, each read out with three probe seeds. All other results come from a single checkpoint per configuration.

\subsection{Evaluation protocol}
\label{sec:evaluation_protocol}

We freeze \mname before every evaluation and consider three complementary protocols. First, single-token and paired-token erasures test whether program edits have localized effects and whether tokens interact beyond their individual contributions (Fig.~\ref{fig:teaser}; Appendix~\ref{fig:pair-synergy-coco-stuff}). Second, code-use statistics and unsupervised region readouts measure vocabulary reuse and the spatial organization of the generated programs (Fig.~\ref{fig:unsupervised-segmentation}). Third, we train lightweight supervised readouts to compare the information retained in frozen DINOv3 features, the interpreted \mname field, and the active quantized program.

The downstream evaluation covers semantic segmentation on PASCAL VOC 2012, COCO-Stuff-27, ADE20K, and Cityscapes; multi-label classification on VOC and COCO-Stuff-27; and monocular depth estimation on NYUv2 (Fig.~\ref{fig:downstream_segmentation_prob_grid} in Appendix~\ref{app:downstream}). For semantic segmentation, all frozen representations are evaluated using the same data pipeline and optimization schedule. The readout architectures differ only where needed to accommodate spatial feature maps and token sequences. Appendix~\ref{app:evaluation_protocol} provides the metric definitions, readout architectures, and complete training and evaluation settings.

\begin{table}[t]
\centering
\caption{Comparison with public adaptive tokenizers at several token budgets. \textbf{Program mIoU} (\%, $\uparrow$) on COCO-Stuff-27 and VOC. Every tokenizer is frozen and read out with the same probe, a transformer with two encoder and two decoder layers over the discrete tokens, trained on the full training split for 40k updates. Bits are nominal index bits per $256{\times}256$ image; for \mname, Tokens is the mean predicted program length and the rate includes the 6-bit length field. Values are means over 3 probe seeds, except rows marked $^\dagger$ (one probe seed). FlexTok~\citep{bachmann2025flextok}, One-D-Piece~\citep{miwa2025onedpiece} and ALIT~\citep{duggal2025alit} are off-the-shelf checkpoints without \mname training, so this is a system-level comparison: training data, teachers and objectives differ. Stuff-27 denotes COCO-Stuff-27. The best value in each column in bold.}
\label{tab:T2}
\setlength{\tabcolsep}{4pt}
\renewcommand{\arraystretch}{1.1}
\begin{tabular*}{\linewidth}{@{\extracolsep{\fill}}llrrrr@{}}
\toprule
Tokenizer & Model & Tokens & Bits & Stuff-27 & VOC \\
\midrule
FlexTok     & D12-D12-IN1k                  & 16  & 256  & 32.7 & 55.6 \\
FlexTok     & D12-D12-IN1k$^\dagger$        & 64  & 1024 & 36.3 & 61.4 \\
FlexTok     & D12-D12-IN1k                  & 256 & 4096 & 35.4 & 61.3 \\
\cmidrule(l){1-6}
One-D-Piece & S-256                         & 21  & 252  & 18.6 & 29.2 \\
One-D-Piece & S-256                         & 256 & 3072 & 21.3 & 31.7 \\
\cmidrule(l){1-6}
ALIT        & small-IN1k                    & 32  & 384  & 16.6 & 23.4 \\
ALIT        & small-IN1k                    & 64  & 768  & 16.5 & 21.8 \\
\midrule
DINO-S      & \msmall (ours), forced prefix$^\dagger$ & 16 & 160 & 35.5 & 56.6 \\
DINO-S      & \msmall (ours), adaptive             & ${\approx}$24.4 & ${\approx}$250 & 39.7 & 62.6 \\
DINO-B      & \msmall (ours), adaptive             & ${\approx}$22.8 & ${\approx}$234 & \textbf{42.3} & \textbf{66.9} \\
\bottomrule
\end{tabular*}
\end{table}

\begin{table}[t]
\centering
\caption{Ablations of the \mname design. Segmentation mIoU (\%) of the adaptive \mname of Table~\ref{tab:T1} (first row) and the change in mIoU (points) when one factor is changed and the model is retrained (DINO-S / \msmall, COCO training). Red/blue: worse/better than \mname, saturating at 10 points. Evaluation as in Table~\ref{tab:T1} (3 tokenizer seeds, each averaged over 3 probe seeds); Mean is the average over the four benchmarks. Stuff-27 denotes COCO-Stuff-27. Measured rates on COCO val are ${\approx}250$ bits per image for \mname, 250 for the spatial grid, ${\approx}248$ for the noncausal generator, ${\approx}244$ for uniform truncation, ${\approx}225$ for $V{=}512$ and ${\approx}276$ for $V{=}2048$; rates of the other variants were not measured. $^\dagger$One of the three seeds collapses (3.0 mean latent mIoU, against 32.5 and 32.8 for the others).  $^\ddagger$2048 implies 11 instead of 10 bit cost (${\approx}276$ vs.\ 250 bits per image); at an equal bit budget, the larger codebook lowers teacher cosine by 0.010 on average, and no seed improves.}
\label{tab:T3}
\definecolor{abneg}{RGB}{178,24,43}\definecolor{abpos}{RGB}{33,102,172}
\setlength{\tabcolsep}{2.5pt}
\renewcommand{\arraystretch}{1.1}
\footnotesize
\begin{tabular*}{\linewidth}{@{\extracolsep{\fill}}l rrrrr rrrrr@{}}
\toprule
 & \multicolumn{5}{c}{Latent} & \multicolumn{5}{c}{Program} \\
\cmidrule(lr){2-6}\cmidrule(lr){7-11}
Variant & VOC & Stuff-27 & ADE20K & Citysc. & Mean & VOC & Stuff-27 & ADE20K & Citysc. & Mean \\
\midrule
\mname (adaptive) & 68.6 & 41.2 & 28.0 & 36.8 & 43.7 & 61.5 & 39.3 & 25.3 & 37.8 & 41.0 \\
\midrule
\emph{Objective} & & & & & & & & & & \\
+ pixel gradients & \cellcolor{abneg!50}$-$10.3 & \cellcolor{abneg!36}$-$5.3 & \cellcolor{abneg!47}$-$8.7 & \cellcolor{abneg!43}$-$7.4 & \cellcolor{abneg!45}$-$7.9 & \cellcolor{abneg!50}$-$11.7 & \cellcolor{abneg!42}$-$7.2 & \cellcolor{abneg!45}$-$8.1 & \cellcolor{abneg!42}$-$7.2 & \cellcolor{abneg!46}$-$8.6 \\
cosine loss only & \cellcolor{abneg!27}$-$2.9 & \cellcolor{abneg!20}$-$1.7 & \cellcolor{abneg!26}$-$2.7 & \cellcolor{abneg!25}$-$2.5 & \cellcolor{abneg!25}$-$2.4 & \cellcolor{abneg!25}$-$2.5 & \cellcolor{abneg!21}$-$1.8 & \cellcolor{abneg!22}$-$1.9 & \cellcolor{abneg!24}$-$2.3 & \cellcolor{abneg!23}$-$2.1 \\
MSE loss only$^\dagger$ & \cellcolor{abneg!50}$-$30.5 & \cellcolor{abneg!50}$-$18.2 & \cellcolor{abneg!50}$-$16.1 & \cellcolor{abneg!50}$-$18.7 & \cellcolor{abneg!50}$-$20.9 & \cellcolor{abneg!50}$-$27.4 & \cellcolor{abneg!50}$-$17.2 & \cellcolor{abneg!50}$-$13.5 & \cellcolor{abneg!50}$-$17.6 & \cellcolor{abneg!50}$-$18.9 \\
\midrule
\emph{Ordering} & & & & & & & & & & \\
noncausal generator & \cellcolor{abneg!30}$-$3.6 & \cellcolor{abneg!19}$-$1.4 & \cellcolor{abneg!25}$-$2.5 & \cellcolor{abneg!29}$-$3.3 & \cellcolor{abneg!26}$-$2.7 & \cellcolor{abneg!27}$-$2.8 & \cellcolor{abneg!20}$-$1.6 & \cellcolor{abneg!22}$-$1.9 & \cellcolor{abneg!30}$-$3.5 & \cellcolor{abneg!25}$-$2.5 \\
spatial $5{\times}5$ grid & \cellcolor{abneg!8}$-$0.2 & \cellcolor{abneg!4}$-$0.1 & \cellcolor{abneg!5}$-$0.1 & 0.0 & \cellcolor{abneg!5}$-$0.1 & 0.0 & \cellcolor{abneg!9}$-$0.3 & \cellcolor{abpos!11}$+$0.5 & \cellcolor{abpos!15}$+$0.9 & \cellcolor{abpos!8}$+$0.3 \\
\midrule
\emph{Curriculum} & & & & & & & & & & \\
joint from start & \cellcolor{abneg!34}$-$4.6 & \cellcolor{abneg!28}$-$3.1 & \cellcolor{abneg!32}$-$4.2 & \cellcolor{abneg!36}$-$5.1 & \cellcolor{abneg!33}$-$4.2 & \cellcolor{abneg!31}$-$3.7 & \cellcolor{abneg!30}$-$3.6 & \cellcolor{abneg!28}$-$3.1 & \cellcolor{abneg!34}$-$4.5 & \cellcolor{abneg!31}$-$3.7 \\
uniform truncation & \cellcolor{abneg!18}$-$1.3 & \cellcolor{abneg!16}$-$1.1 & \cellcolor{abneg!20}$-$1.6 & \cellcolor{abneg!20}$-$1.7 & \cellcolor{abneg!19}$-$1.4 & \cellcolor{abneg!12}$-$0.6 & \cellcolor{abneg!16}$-$1.0 & \cellcolor{abneg!14}$-$0.8 & \cellcolor{abneg!21}$-$1.8 & \cellcolor{abneg!16}$-$1.1 \\
abrupt handoff & \cellcolor{abpos!6}$+$0.1 & 0.0 & \cellcolor{abneg!6}$-$0.2 & \cellcolor{abneg!11}$-$0.5 & \cellcolor{abneg!6}$-$0.1 & \cellcolor{abneg!4}$-$0.1 & 0.0 & \cellcolor{abpos!8}$+$0.2 & \cellcolor{abneg!6}$-$0.1 & 0.0 \\
no oracle warm-up & \cellcolor{abpos!4}$+$0.1 & \cellcolor{abpos!7}$+$0.2 & 0.0 & 0.0 & \cellcolor{abpos!4}$+$0.1 & \cellcolor{abneg!9}$-$0.3 & \cellcolor{abpos!5}$+$0.1 & \cellcolor{abpos!13}$+$0.7 & \cellcolor{abneg!11}$-$0.5 & 0.0 \\
\midrule
\emph{Vocabulary} & & & & & & & & & & \\
$V{=}512$ & \cellcolor{abneg!21}$-$1.7 & \cellcolor{abneg!17}$-$1.1 & \cellcolor{abneg!20}$-$1.5 & \cellcolor{abneg!23}$-$2.2 & \cellcolor{abneg!20}$-$1.6 & \cellcolor{abneg!19}$-$1.5 & \cellcolor{abneg!16}$-$1.1 & \cellcolor{abneg!18}$-$1.3 & \cellcolor{abneg!21}$-$1.8 & \cellcolor{abneg!19}$-$1.4 \\
$V{=}2048^\ddagger$ & \cellcolor{abpos!14}$+$0.8 & \cellcolor{abpos!12}$+$0.5 & \cellcolor{abpos!16}$+$1.1 & \cellcolor{abpos!22}$+$1.9 & \cellcolor{abpos!16}$+$1.1 & \cellcolor{abpos!10}$+$0.4 & \cellcolor{abpos!13}$+$0.7 & \cellcolor{abpos!17}$+$1.2 & \cellcolor{abpos!19}$+$1.4 & \cellcolor{abpos!15}$+$0.9 \\
\bottomrule
\end{tabular*}
\end{table}

\subsection{Adaptive versus fixed-length programs}
\label{sec:rate-matched}
Table~\ref{tab:T1} compares \mname with fixed-length tokenizers trained separately with the same encoder, generator, codebook, and data. At about $250$ bits per image, \mname beats the fixed $K{=}25$ model on all four benchmarks and for both readouts, by $1.6$--$3.1$ percent points on mIoU. It also beats the fixed $K{=}32$ model everywhere with $22\%$ fewer bits. Only the fixed $K{=}64$ model scores higher, at $2.6{\times}$ the rate. Forcing the same adaptive checkpoint to longer prefixes does not help: its predicted length matches or exceeds every forced length up to $64$ codes (Fig.~\ref{fig:rate-quality-coco}). At a fixed average budget, the improvements therefore come from spending more codes on some images and fewer on others.

\textbf{What drives the gain.}
Table~\ref{tab:T3} retrains the model with one factor changed at a time. Letting pixel-reconstruction gradients reach the generator costs the most, $7.9$ mIoU for the latent readout and $8.6$ for the program readout on average. Training all curriculum phases jointly from the start costs $4.2$/$3.7$ points and a noncausal generator $2.7$/$2.5$. Sampling Phase-1 truncation lengths uniformly, with no bias toward long prefixes, costs $1.4$/$1.1$, while the handoff style and the oracle warm-up change little. The latent loss needs both of its terms: cosine alone loses $2.4$/$2.1$ points, and with MSE alone one of three seeds collapses. A $5{\times}5$ spatial grid with the same $25$ codes performs on par with \mname, so we do not claim that the sequential layout is better in itself. At a fixed length of $25$ codes the sequential model trails that grid by $2.4$/$2.2$ points (Table~\ref{tab:ablation-absolute}), and adaptive allocation closes the difference. The main contrasts also hold with ImageNet training, with a DINO-B teacher, and with the larger \mmedium generator, although the curriculum matters less for \mmedium (Appendix~\ref{app:replication}).

\subsection{Comparison with public adaptive tokenizers}
\label{sec:external}
Table~\ref{tab:T2} applies the transformer readout used for \mname programs to frozen tokens from FlexTok~\citep{bachmann2025flextok}, One-D-Piece~\citep{miwa2025onedpiece}, and ALIT~\citep{duggal2025alit}. At about $250$ bits per image, \mname reaches $39.7$ mIoU on COCO-Stuff-27 and $62.6$ on VOC, $7.0$ points above FlexTok at $256$ bits. With a DINO-B teacher it reaches $42.3$ and $66.9$ at about $234$ bits. Longer sequences help the public tokenizers little. FlexTok peaks at $36.3$ on COCO-Stuff-27 with $1024$ bits and at $61.9$ on VOC with $2048$ bits, and One-D-Piece and ALIT stay below $22$ mIoU on COCO-Stuff-27 and $32$ on VOC at every budget (Fig.~\ref{fig:rate-quality-external}). The tokenizers differ in training data, teacher, and objective, so the comparison measures how much segmentation-relevant content each system keeps per bit and cannot attribute the difference to a single component.

\subsection{Alignment, transfer and structure}
\label{sec:downstream}
\textbf{Alignment.} Under the same 640-bit ceiling, alignment with the teacher falls as the teacher grows (cosine $0.86$ for DINO-S, $0.81$ for DINO-B and $0.75$ for DINO-L with \msmall), yet DINO-B and DINO-L segment better than DINO-S (Appendix~\ref{app:alignment}).
\textbf{Downstream.} Across all ten variants, the COCO-trained DINO-B/\msmall model segments best, with $47.3$ mean latent mIoU against $61.3$ for its frozen teacher. A small transformer readout on the program trails the latent readout by $2$--$7$ mIoU on VOC, COCO-Stuff-27 and ADE20K, is ahead on Cityscapes, and reaches a slightly lower RMSE on NYUv2 depth, so most of the dense-prediction information survives in the discrete program (Appendix~\ref{app:downstream}).
\textbf{Structure.} Erasing a single token changes a spatially localized part of the interpreted field (Fig.~\ref{fig:teaser}; token pairs in Appendix~\ref{app:pairwise-token-synergies}), and clustering these erasure effects gives image regions similar to those obtained from DINO and latent features (Fig.~\ref{fig:unsupervised-segmentation}). Program length correlates with object count on CLEVR (Pearson $r \approx 0.74$) and more weakly with scene complexity on COCO.

%% file: body/limitations.tex
\section{Limitations}

(i) The most visible limitation is the readability of the raw program. With a small transformer readout, the program comes close to the interpreted field on depth and on most segmentation benchmarks. On multi-label classification, where the original attention-pooled linear readout is used, it trails far behind (Appendix~\ref{app:downstream}). %
Separating readout capacity from information lost at quantization will need readouts designed for programs that make fuller use of token order and prefix structure.
(ii) Latent alignment is an incomplete predictor of transfer. The DINO-S
encoder gives the best reconstruction scores in Table~\ref{tab:test-metrics},
but DINO-B and DINO-L transfer better (Table~\ref{tab:T5b} in Appendix~\ref{app:alignment}).
Useful structure in
the teacher features can therefore survive quantization even when the raw
reconstruction loss is higher. This calls for more discriminative latent-quality measures. 
(iii) We compare public adaptive
tokenizers on COCO-Stuff-27 and VOC only, with a single frozen-token readout
(Table~\ref{tab:T2}). The systems differ in training data, teacher, and
objective, so this comparison does not isolate adaptive length; Table~\ref{tab:T1}
tests that against separately trained fixed-length models.
(iv) The evidence is also unevenly replicated. The rate-matched comparison and
the ablations use three tokenizer seeds, and their main contrasts hold for
other training data, teachers, and generator sizes
(Appendix~\ref{app:replication}). The model-size grid uses one checkpoint
per configuration, and the rate-matched fixed baseline is calibrated on COCO.
(v) LLM scene decoding (Appendix~\ref{app:llm_scene_decoding}) recovers
coarse layout but not reliable attribute binding. The learned codebook is
therefore not yet a fully transparent symbolic interface for downstream
language models. Closing this gap, either by changing the codebook
structure or by aligning code identities with a captioning signal, is a
direction we leave to future work.

%% file: ack.tex
\begin{ack}
PW is funded by the statutory funds of Poznan University of Technology and the Polish Ministry of Science and Higher Education, grant no. 0311/SBAD/0774 and Research Grant of National Science Center, grant no. 2024/53/N/ST6/03961. KD and KK are funded by the statutory funds of Poznan University of Technology and the Polish Ministry of Science and Higher Education, grant no. 0311/SBAD/0770 and the Research Grant of National Science Center, grant no. 2025/57/B/ST6/03737. We gratefully acknowledge the Polish high-performance computing infrastructure PCSS PLCloud for providing computational resources and support under grant no. pl0603-01.
\end{ack}

%% file: appendix.tex
\section{Adaptive program-length curriculum details}
\label{app:curriculum-details}

This appendix gives the full parameterization of the adaptive-length curriculum used in Section~\ref{sec:curriculum}. 

The generator emits a fixed sequence of $K$ quantized tokens for every image, but the interpreter attends only to a prefix of length $L_i \in \{1,\ldots,K\}$ for sample $x_i$. Tokens with positions $k > L_i$ are masked during training; at inference they are not passed to the interpreter. %
Curriculum design therefore reduces to the prediction of $L_i$ during training, which we schedule across the following four phases. %

\textbf{Phase 1: Random truncation.}
The model is trained only on randomly truncated programs, which forces
informative content to be front-loaded into earlier tokens. Lengths are
sampled as
\begin{equation}
u_i \sim \mathrm{Beta}(\alpha_t, 1),
\qquad
L_i = \mathrm{clip}\!\left(\mathrm{round}\!\left(L_{\min}^{\mathrm{trunc}}
+ u_i (K - L_{\min}^{\mathrm{trunc}})\right),\,
L_{\min}^{\mathrm{trunc}},\, K\right),\label{eq:rand-trunc}
\end{equation}
with the schedule
$\alpha_s = \alpha_0 + (1-\alpha_0)\min(s/T_{\mathrm{bias}}, 1)$, where $s$ denotes the global optimization step.
Since $\alpha_0 > 1$, early training is biased toward longer prefixes
(easier reconstruction, faster initial convergence) and anneals linearly
to a uniform distribution at
$s = T_{\mathrm{bias}}$.

\textbf{Phase 2: Oracle target estimation.}
We continue with random truncation but begin %
estimating for each sample how long the program
\emph{should} be. The base oracle uses the 
feature-space reconstruction error at the active length,
$e_i(L_i)=\mathcal{L}_{\mathrm{lat}}(Z_{\mathrm{out},i}(L_i),Z_{e,i})$,
normalized by a batch-level EMA
$\bar{e}_t = \rho\,\bar{e}_{t-1} + (1-\rho)\,\tfrac{1}{B}\sum_i e_i(L_i)$:
\begin{equation}
L_i^{\mathrm{base}}
= \mathrm{clip}\!\left(
\tfrac{e_i(L_i)}{\bar{e}_t + \epsilon}\,\beta K,\;
L_{\min}^{\mathrm{oracle}},\, L_{\max}^{\mathrm{oracle}}\right).
\end{equation}
Samples reconstructing worse than the EMA running average receive longer
targets; easier samples receive shorter ones. The hyperparameter $\beta$
sets the average target length when $e_i \approx \bar{e}_t$. Targets are
not yet used to drive truncation in this phase—they are only accumulated.

As $L_i^{\mathrm{base}}$ conflates sample difficulty with length adequacy (a hard
sample has large error at any length), we complement it with a signal of the local slope of the rate--feature-distortion curve at $L_i$,
which captures how much the \emph{additional} tokens reduce
error. For a probe radius $\delta$, we calculate the reconstruction error at the clipped
neighbors
$L_i^{\pm} = \mathrm{clip}(L_i \pm \delta,\, L_{\min}^{\mathrm{trunc}},\, K)$
and form one-sided relative slopes
\begin{equation}
r_i^{\mathrm{short}}
= \max\!\left(0,\, \tfrac{e_i^- - e_i}{e_i+\epsilon}\right),
\qquad
r_i^{\mathrm{long}}
= \max\!\left(0,\, \tfrac{e_i - e_i^+}{e_i+\epsilon}\right),
\end{equation}
which measure how much shortening hurts and how much lengthening still
helps. We average over non-clamped probes, maintain EMAs
$\bar{r}^{\mathrm{short}}, \bar{r}^{\mathrm{long}}$, and squash with a
temperature-$\tau$ saturation
$u_\bullet = \tanh(\bar{r}^\bullet / \tau) \in [0,1]$. The pair
$(u_{\mathrm{short}}, u_{\mathrm{long}})$ induces a soft three-way decision
through the partition of unity
\begin{equation}
w_{\mathrm{compress}} = (1-u_{\mathrm{short}})(1-u_{\mathrm{long}}),
\quad
w_{\mathrm{keep}} = u_{\mathrm{short}}(1-u_{\mathrm{long}}),
\quad
w_{\mathrm{extend}} = u_{\mathrm{long}},
\end{equation}
corresponding to three regimes: \emph{compress} when the curve is flat
near $L_i$, \emph{keep} near the elbow (only shortening hurts), and
\emph{extend} whenever lengthening still pays. %
Extension dominates by design—unused marginal utility should be exploited
regardless of the lower-side signal. The corrected target is
\begin{equation}
\tilde{L}_i
= \mathrm{clip}\!\left( m\, L_i^{\mathrm{base}},\;
L_{\min}^{\mathrm{oracle}},\, L_{\max}^{\mathrm{oracle}}\right),
\qquad
m = \sum_k m_k\, w_k,
\end{equation}
with defaults $m_{\mathrm{compress}} < 1$, $m_{\mathrm{keep}} = 1$, and
$m_{\mathrm{extend}} > 1$.

\textbf{Phase 3: Supervised length-head training.}
We activate a length head
$\hat{L}_i = K h_\phi(\mathrm{pool}(Z_{e,i}))$,
a sigmoid MLP on pooled DINOv3 encoder features, supervised against the
oracle target:
\begin{equation}
\mathcal{L}_{\mathrm{len}}
= \tfrac{1}{B}\sum_i
\left(\tfrac{\hat{L}_i - \mathrm{sg}[\tilde{L}_i]}{K}\right)^2 .
\end{equation}
The objective $\mathcal{L}_{\mathrm{len}}$ is optimized only with respect to the head parameters $\phi$ from this point onward. During Phase 3, the head is being trained but not yet deployed: truncation is still random (Eq.\ \ref{eq:rand-trunc}), so $h_\phi$ learns from a stable target distribution before controlling the active prefix.

\textbf{Phase 4: Handoff to predicted lengths.}
Let $s^{(4)}_{\mathrm{ramp,start}}$ and
$s^{(4)}_{\mathrm{ramp,end}}$ be the first and last steps of the handoff ramp.
We linearly ramp
$a_s =
\mathrm{clip}(
(s-s^{(4)}_{\mathrm{start}})/
(s^{(4)}_{\mathrm{end}}-s^{(4)}_{\mathrm{start}}),
0, 1
)$
and truncate the program by sampling based on the predicted length $\hat{L}_i$ as
\begin{equation}
L_i =
\begin{cases}
\mathrm{clip}(\mathrm{round}(\hat{L}_i), 1, K) & \text{w.p. } a_s,\\
L_i^{\mathrm{rand}} & \text{w.p. } 1 - a_s,
\end{cases}
\end{equation}
Thus, Phase 4 gradually replaces random truncations with predicted ones, thereby avoiding the covariate shift of an abrupt switch. After $s \ge s^{(4)}_{\mathrm{end}}$, the ramp has completed: all remaining training steps use predicted truncation lengths, while the length head continues to be updated by $\mathcal{L}_{\mathrm{len}}$. The head continues to be trained with $\mathcal{L}_{\mathrm{len}}$ throughout this phase. 

At inference, no sampling or length loss is used: $\hat{L}_i$ is rounded and clipped to $[1,K]$, and the program is hard-truncated at that length before decoding.

\begin{table}[t]
\centering
\small
\caption{Curriculum phase schedule (in training steps $\times 10^3$) and shared hyperparameters. \mname-S variants begin the curriculum earlier to match their faster convergence; all other settings are identical across variants and datasets.}
\label{tab:curriculum}
\begin{tabular}{lcc}
\toprule
& \mname-S & \mname-M \\
\midrule
Phase 1: random truncation only  & 0--100k & 0--200k \\
Phase 2: + oracle targets        & 100k--150k & 200k--250k \\
Phase 3: + length-head training  & 150k--200k & 250k--300k \\
Phase 4: handoff ramp            & 200k--250k & 300k--350k \\
Predicted lengths only           & 250k--500k & 350k--500k \\
\midrule
\multicolumn{3}{l}{\textit{Shared curriculum hyperparameters}} \\
\midrule
Oracle: $\beta{=}0.75$, $\rho{=}0.999$, $L_{\min}^{\mathrm{oracle}}{=}2$ & \multicolumn{2}{c}{} \\
Probes: $\delta{=}2$, $\tau{=}0.3$, slope EMA${=}0.99$ & \multicolumn{2}{c}{} \\
Modulators: $m_{\mathrm{compress}}{=}0.4$, $m_{\mathrm{keep}}{=}1.0$, $m_{\mathrm{extend}}{=}1.3$ & \multicolumn{2}{c}{} \\
Truncation: $L_{\min}^{\mathrm{trunc}}{=}8$,\; $\lambda_{\mathrm{len}}{=}1.0$ & \multicolumn{2}{c}{} \\
\bottomrule
\end{tabular}
\end{table}

Table~\ref{tab:curriculum} lists the phase schedule used for all main variants. The only difference between \mname-S and \mname-M is the onset of Phase~2: smaller generators converge faster during random truncation, so the curriculum begins at 100k steps rather than 200k. Each subsequent phase occupies a 50k-step window. All oracle, probe, and modulator hyperparameters are shared.

\begin{table}[t]
\centering
\small
\caption{Training objective, quantizer and optimization settings shared across all model variants.}
\label{tab:training}
\begin{tabular}{ll}
\toprule
\textbf{Component} & \textbf{Configuration} \\
\midrule
Program    & $K{=}64$ tokens, 2D patch + learned prefix positional encoding \\
Quantizer  & $V{=}1024$, $d_c{=}16$, $\ell_2$-normalized, EMA ($\gamma{=}0.95$), $\beta{=}1.0$ \\
Decoder    & FastGAN, L1 + 0.1 LPIPS-VGG, trained with stop-gradient on interpreter output \\
\midrule
$\mathcal{L}_{\text{lat}}$     & cosine + MSE vs.\ frozen DINOv3 patches, $\lambda_{\text{lat}}{=}1.0$ \\
$\mathcal{L}_{\text{commit}}$    & commitment loss, $\lambda_{\text{q}}{=}1.0$ \\
$\mathcal{L}_{\text{div}}$       & diversity regularizer, $\lambda_{\text{div}}{=}0.3$, 40k-step warmup \\
\midrule
Optimizer  & AdamW, batch size 32, $5{\times}10^{5}$ steps \\
Schedule   & peak LR $10^{-3}$; 30k-step warmup from $0.1{\times}$, hold to 100k, cosine decay to $10^{-4}$ at 500k \\
Length head & learning rate $3{\times}10^{-4}$ \\
\bottomrule
\end{tabular}
\end{table}

\FloatBarrier
\section{Curriculum timing stability}
\label{app:stability}

We ablated the \emph{timing} of the phases rather than their removal: both schedules run the same four phases followed by training entirely on predicted lengths, under the same $500$k-step budget. The schedule used in the main experiments completes the handoff by step $50$k; the prolonged schedules complete it between steps $250$k and $450$k depending on generator size. Across ten otherwise matched configurations, the prolonged schedule reduced the final full-capacity composite validation loss in $8/10$ cases. This motivated the final schedule and shows that a large change to when the handoff happens does not systematically destabilize training.

\FloatBarrier
\section{Qualitative program lengths}
\label{app:qualitative-lengths}

To give a direct sense of what the length head does on individual images, we
show representative COCO crops selected along two axes: the predicted length
itself, and the gap between the predicted length and what the reconstruction
curve actually needs. For the latter we define $L^{*}_{95}$ as the shortest
prefix length at which reconstruction quality reaches $95\%$ of the quality at
full length $K{=}64$, computed per image from its own rate--distortion curve.
Each panel caption reports the predicted length $L_{\mathrm{pred}}$,
$L^{*}_{95}$, reconstruction RMSE at $L_{\mathrm{pred}}$, the gap to full
length $\Delta_{64} = \mathrm{RMSE}(L_{\mathrm{pred}}) - \mathrm{RMSE}(K{=}64)$,
and the number of annotated classes present in the crop.

\begin{figure}[t]
    \centering
    \includegraphics[width=\linewidth]{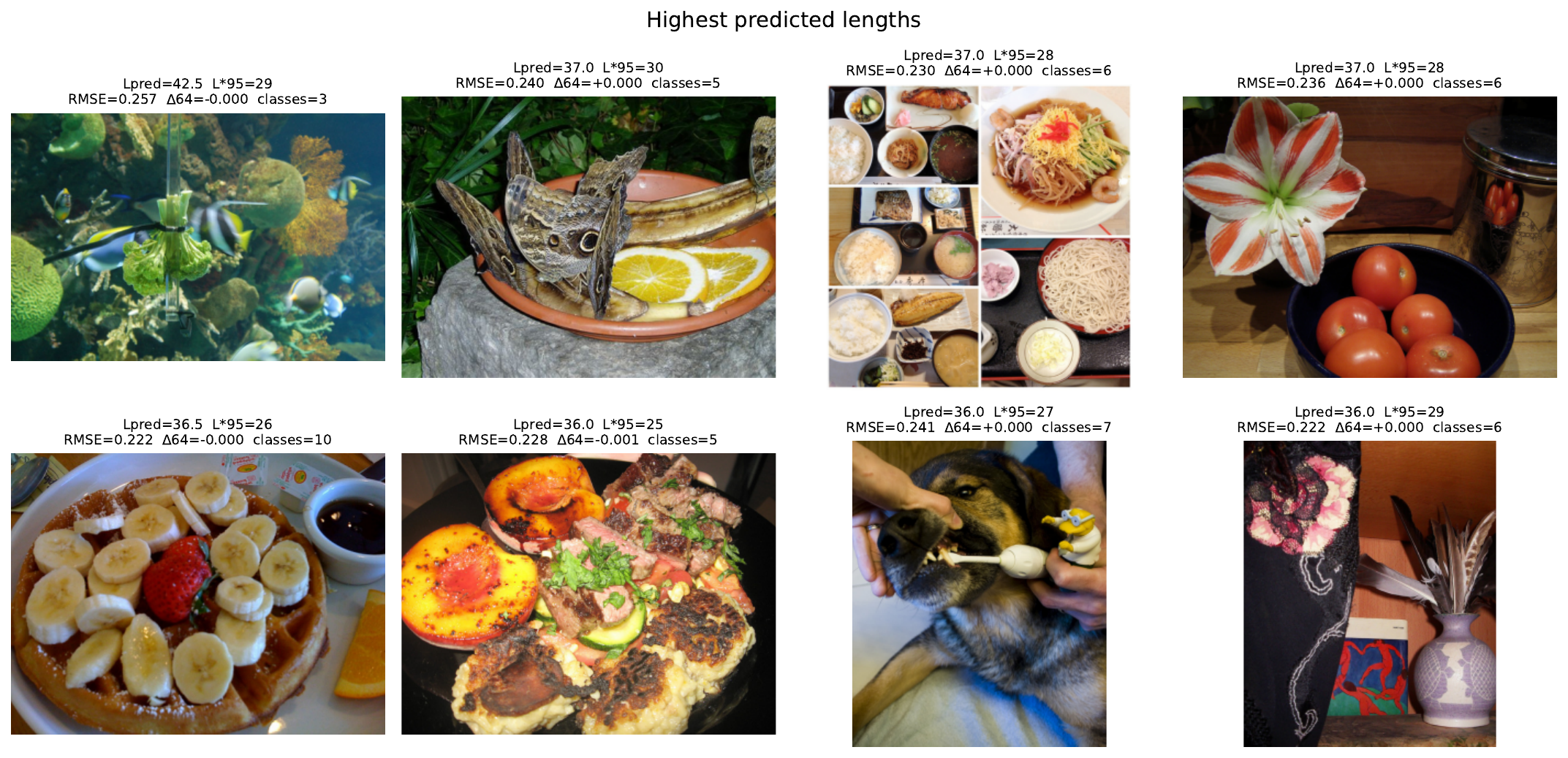}
    \caption{\textbf{Longest predicted programs.} Crops receiving the highest
    predicted lengths $L_{\mathrm{pred}}$, typically cluttered or multi-object
    scenes.}
    \label{fig:qual-longest}
\end{figure}

\begin{figure}[t]
    \centering
    \includegraphics[width=\linewidth]{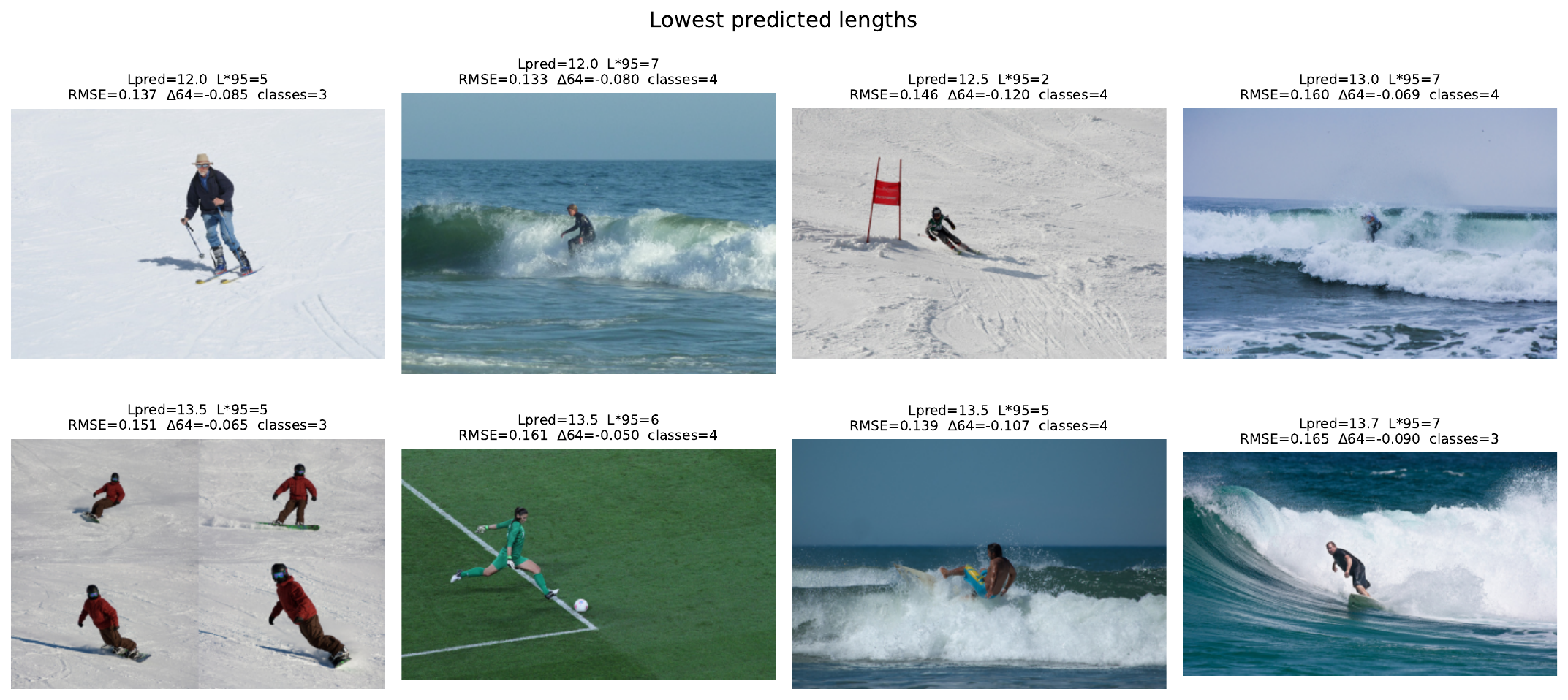}
    \caption{\textbf{Shortest predicted programs.} Crops receiving the lowest
    predicted lengths $L_{\mathrm{pred}}$, typically low-clutter,
    single-subject scenes.}
    \label{fig:qual-shortest}
\end{figure}

\begin{figure}[t]
    \centering
    \includegraphics[width=\linewidth]{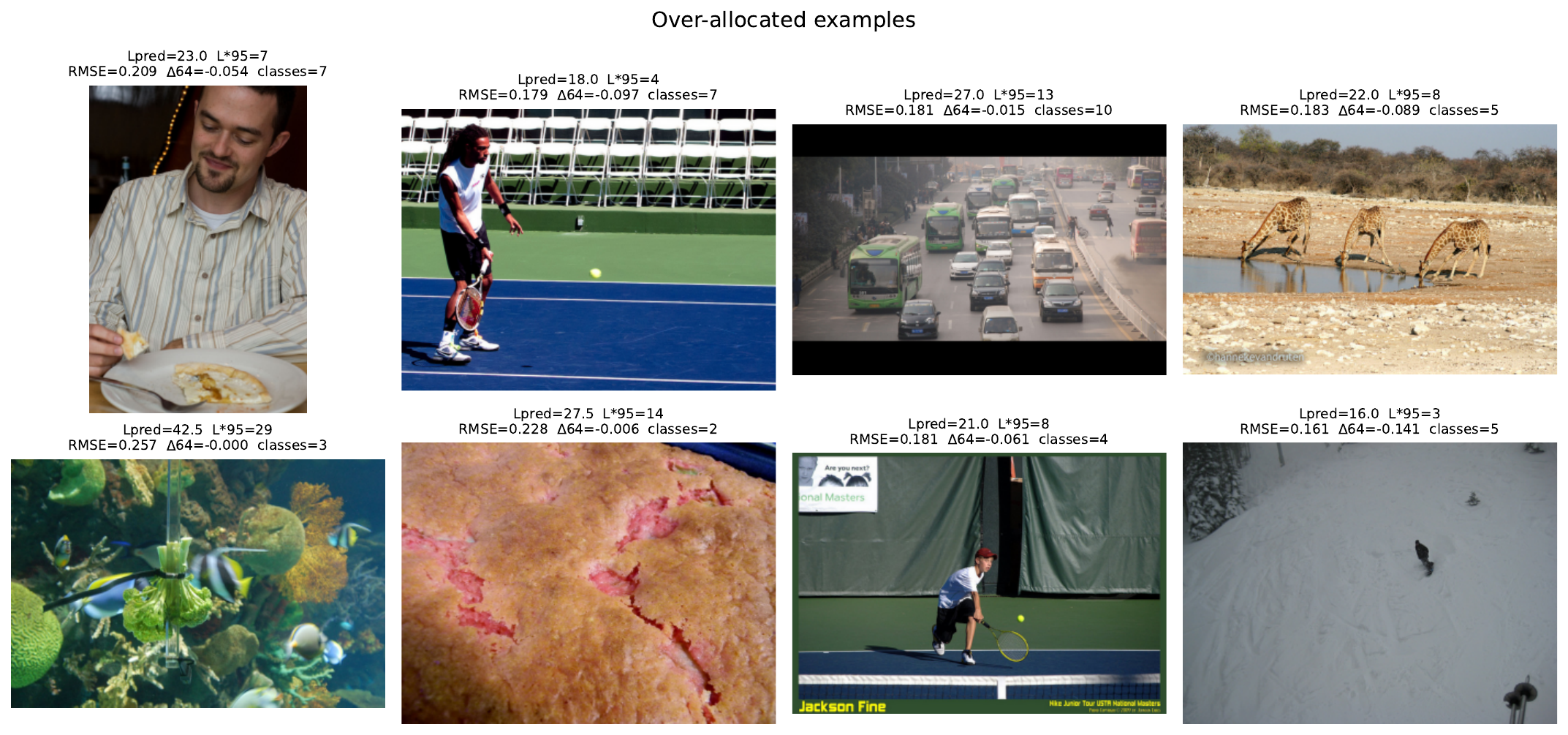}
    \caption{\textbf{Over-allocated.} Crops where $L_{\mathrm{pred}}$ exceeds
    $L^{*}_{95}$ by the largest margin: the head spends more tokens than the
    rate--distortion curve indicates are needed.}
    \label{fig:qual-over}
\end{figure}

\begin{figure}[t]
    \centering
    \includegraphics[width=\linewidth]{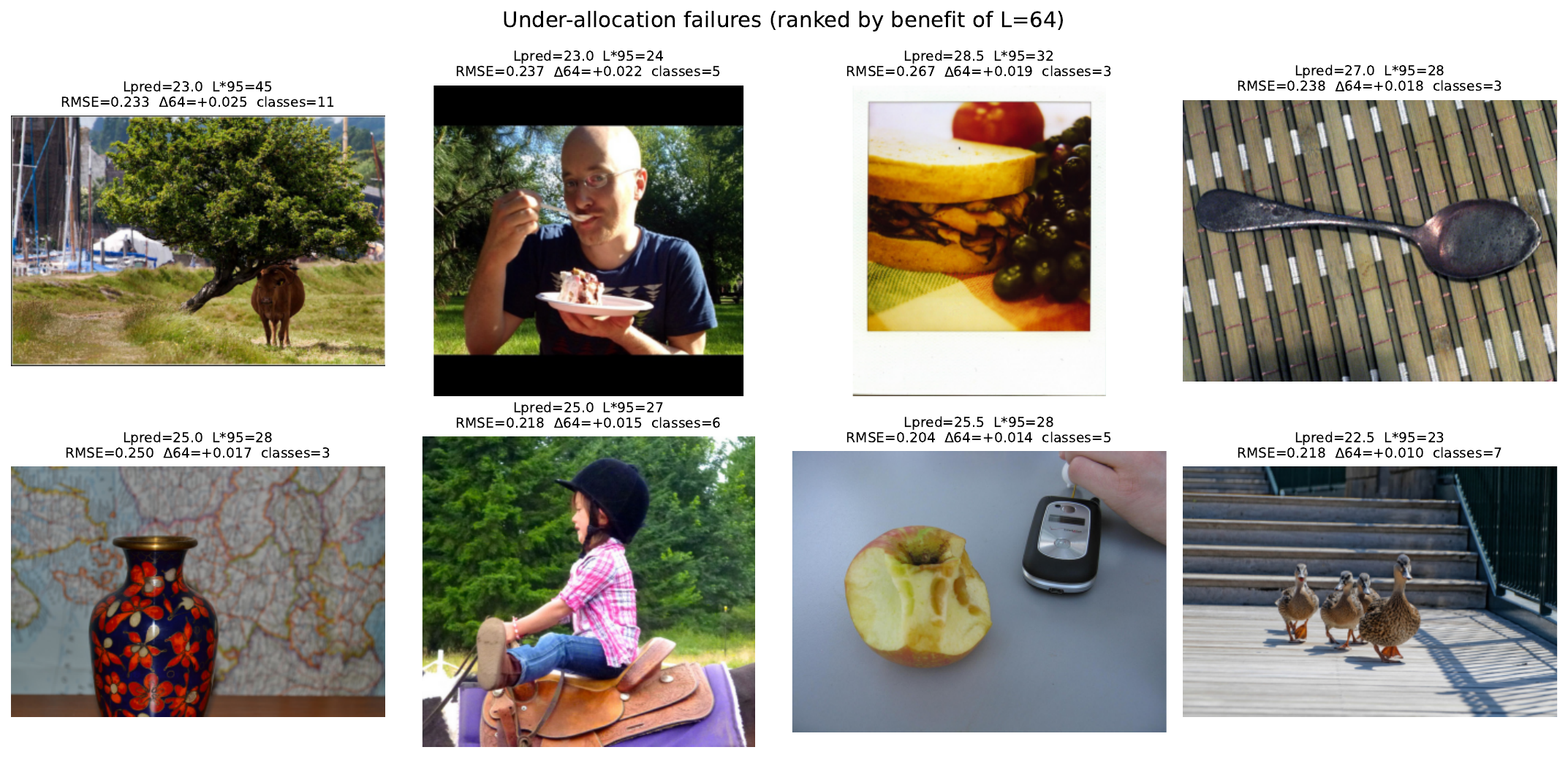}
    \caption{\textbf{Under-allocated.} Crops where $L^{*}_{95}$ exceeds
    $L_{\mathrm{pred}}$ by the largest margin: the head spends fewer tokens
    than the curve indicates are needed.}
    \label{fig:qual-under}
\end{figure}

Figure~\ref{fig:qual-longest} and Figure~\ref{fig:qual-shortest} show the
programs \mname allocates the most and the fewest tokens to. Figures
\ref{fig:qual-over} and~\ref{fig:qual-under} isolate cases where
$L_{\mathrm{pred}}$ and $L^{*}_{95}$ disagree: \emph{over-allocated} crops
receive more tokens than the curve indicates are needed, and
\emph{under-allocated} crops receive fewer. These disagreement cases are the
more informative ones, since they show where the length head's estimate and
the direct rate--distortion measurement diverge, rather than where they agree.

\FloatBarrier
\section{Architectural details}\label{app:variants}
Table~\ref{tab:variants} lists the STROP variants used in the main experiments, including backbone, generator--interpreter size, parameter count, and single-H100 training time.

\begin{table}[h]
\centering
\small
\caption{Model variants trained on both COCO and ImageNet. The interpreter mirrors the generator's architecture. Training time is wall-clock on a single NVIDIA H100 (ImageNet).}
\label{tab:variants}
\begin{tabular}{lccccccc}
\toprule
Encoder & Model & $d_{\text{model}}$ & Layers & FFN & \makecell{Trainable (M)} & \makecell{Total (M)} & \makecell{Time (h)} \\
\midrule
DINO-S & \msmall & 256  & 8  & 1024 & 13.5 & 35.1  & 27 \\
DINO-S & \mmedium & 512  & 10 & 2048 & 64.3 & 85.9  & 45 \\
DINO-B & \msmall & 256  & 8  & 1024 & 14.2 & 99.8  & 32 \\
DINO-B & \mmedium & 512  & 10 & 2048 & 65.1 & 150.8 & 50 \\
DINO-L & \mmedium & 512  & 10 & 2048 & 65.6 & 368.8 & 66 \\
\bottomrule
\end{tabular}
\end{table}
\FloatBarrier

\section{Evaluation protocol details}
\label{app:evaluation_protocol}

\textbf{Frozen representations.}
All representation models are frozen before evaluation. We train a separate readout for each representation and dataset. Gradients from the readouts do not update DINOv3, the programmer, the codebook, or the interpreter. Unless stated otherwise, each result uses one probe initialization. The rate-matched comparison and the ablations (Tables~\ref{tab:T1} and~\ref{tab:T3}) average three probe seeds for each of three tokenizer seeds; the external-tokenizer comparison uses three probe seeds where marked (Fig.~\ref{fig:rate-quality-external}).

\textbf{Token-erasure interventions.}
Let $Z_{\mathrm{out}}$ denote the interpreted patch field obtained from the complete active program. To measure the effect of token position $k$, we mask that position and run the interpreter again, obtaining $Z_{\mathrm{out}}^{-k}$. Masking preserves the original sequence length and token indices. The patchwise erasure effect is
\begin{equation}
    \Delta_k(s)
    =
    1-\cos\left(
        Z_{\mathrm{out}}(s),
        Z_{\mathrm{out}}^{-k}(s)
    \right),
\end{equation}
where $s$ indexes a spatial position. We assign each patch to the active token whose removal produces the largest change,
\begin{equation}
    a(s)
    =
    \arg\max_{k\in\mathcal{A}}\Delta_k(s),
\end{equation}
where $\mathcal{A}$ is the set of active token positions. Masking inactive positions provides a control for the single-token intervention.

For a pair of active positions $(i,j)$, let $Z_{\mathrm{out}}^{-ij}$ denote the interpreted field obtained after masking both tokens. We define
\begin{equation}
    \Delta_{ij}(s)
    =
    1-\cos\left(
        Z_{\mathrm{out}}(s),
        Z_{\mathrm{out}}^{-ij}(s)
    \right)
\end{equation}
and measure their positive supra-additive interaction as
\begin{equation}
    \operatorname{Syn}^{+}_{ij}(s)
    =
    \max\left(
        \Delta_{ij}(s)-\Delta_i(s)-\Delta_j(s),
        0
    \right).
\end{equation}
The scalar pair score is the mean over the $10\%$ of patches with the largest positive interaction values. At the $16\times16$ feature resolution, this corresponds to 26 patches. These interventions are used in Figs.~\ref{fig:teaser} and~\ref{fig:pair-synergy-coco-stuff}. The sampling and qualitative selection procedure for Fig.~\ref{fig:pair-synergy-coco-stuff} is described alongside that figure.

\textbf{Vocabulary reuse.}
Let $V$ be the vocabulary size, $n_c$ the number of occurrences of code $c$, and
\begin{equation}
    N_{\mathrm{tok}}=\sum_c n_c
\end{equation}
the total number of active program tokens. We report the fraction of vocabulary entries used at least once,
\begin{equation}
    \mathrm{V}\%
    =
    100\,
    \frac{
        \left|\left\{c:n_c>0\right\}\right|
    }{V}.
\end{equation}

To measure how evenly the vocabulary is used, we define $\pi_c=n_c/N_{\mathrm{tok}}$ and compute the empirical code entropy
\begin{equation}
    \mathcal{H}_{\mathrm{code}}
    =
    -\sum_{c:n_c>0}\pi_c\log\pi_c.
\end{equation}
The effective vocabulary size is $\exp(\mathcal{H}_{\mathrm{code}})$. We report it relative to the complete vocabulary as
\begin{equation}
    \mathrm{Eff.}\%
    =
    100\,
    \frac{\exp(\mathcal{H}_{\mathrm{code}})}{V}.
\end{equation}
A high $\mathrm{V}\%$ indicates broad vocabulary coverage. The effective utilization remains low when most occurrences are concentrated on a small subset of codes.

\textbf{Unsupervised region evaluation.}
The results in Fig.~\ref{fig:unsupervised-segmentation} use bundles of 512 images. Images 0--255 form the calibration subset, while images 256--511 are held out for evaluation. The calibration images come from the validation bundle; the official training split is unused.

For the DINOv3 and interpreted-field readouts, we cluster patch vectors from $Z_e$ and $Z_{\mathrm{out}}$, respectively. Both are $\ell_2$-normalized before clustering. For the token readout, we rank active positions by their image-averaged erasure effect and retain at most 16 positions. At each spatial position $s$, their erasure values form the attribution vector
\begin{equation}
    \boldsymbol{\Delta}(s)
    =
    \left(
        \Delta_{k_1}(s),
        \ldots,
        \Delta_{k_{16}}(s)
    \right).
\end{equation}
Vectors from programs with fewer than 16 active positions are padded with zeros. Padded code identifiers are set to $-1$. The attribution vectors are not normalized before clustering.

We fit $k$-means using at most 40k labeled calibration patches, 40 iterations, one initialization, and seed 0. We use 27 clusters for COCO-Stuff-27 and 11 clusters for CLEVR.

For code-to-class transfer, each patch is assigned to the active token with the largest erasure effect. We then associate the code at that token position with its most frequent ground-truth class on the calibration subset. Void pixels are excluded when estimating this mapping. Codes absent from the calibration subset predict the void class. The resulting mapping remains fixed during evaluation on the held-out images.

\textbf{Token grounding.}
For an image, we rank the active program tokens by their image-averaged erasure effect and keep the $k$ most influential. Each token's map $\Delta_k$ is clipped at zero, divided by its 99th percentile and clipped to $[0,1]$; patches with a normalized value of at least $0.45$ form the token's predicted mask. Ground-truth object masks are projected to the $16{\times}16$ patch grid, and a patch belongs to an object when the object covers at least $20\%$ of it. Predicted and ground-truth masks are matched one-to-one by the Hungarian algorithm on their IoU. We report the mean IoU over ground-truth objects (unmatched objects count as zero), the fraction of objects with IoU $\geq 0.5$ (Recall@0.5), and the mean IoU over matched pairs. Scores are averaged over images; the comparison between models is relative, since no random-token baseline is reported.

\textbf{Semantic-segmentation readouts.}
The semantic-segmentation results in Fig.~\ref{fig:downstream_segmentation_prob_grid} use the complete official training and validation splits. Spatial readouts operate on $Z_e$ or $Z_{\mathrm{out}}$. The program readout receives the active prefix of $Z_q$.

During training, we sample a scale $r\sim\mathcal{U}(0.5,2)$ and resize the image so that its shorter side is $\lfloor256r\rfloor$, preserving its aspect ratio. Images use bilinear interpolation with antialiasing, while segmentation masks use nearest-neighbor interpolation. We then extract a random $256\times256$ crop. When a multi-class crop is possible, sampling is repeated up to ten times if one non-void class occupies at least $75\%$ of the labeled pixels. Horizontal flipping is applied with probability $0.5$.

The photometric augmentation follows the DINOv3 evaluation pipeline. Independently with probability $0.5$, we apply a brightness offset sampled from $[-32,32]$, a contrast factor from $[0.5,1.5]$, a saturation factor from $[0.5,1.5]$, and a hue shift from $[-0.5,0.5]$. Contrast is placed randomly before or after the saturation and hue transformations. Images are normalized using the ImageNet mean $(0.485,0.456,0.406)$ and standard deviation $(0.229,0.224,0.225)$. If padding is required, normalized images are padded with zeros and target masks with the ignore label 255. The frozen representations are computed online from the augmented images.

The readout for $Z_e$ and $Z_{\mathrm{out}}$ consists of \texttt{Dropout2d} with probability $0.1$, synchronized batch normalization, and a $1\times1$ convolution. For $Z_q$, the active quantized vectors are projected to width 256 and combined with learned positional embeddings. The program readout contains two pre-normalization Transformer encoder layers, 256 learned output queries arranged as a $16\times16$ field, and two Transformer decoder layers. Each layer uses four attention heads, an FFN width of 1024, and dropout $0.1$. A final linear projection produces the class logits.

We train each readout for 40k updates, including 1.5k warmup updates, using AdamW with a learning rate of $10^{-3}$. The global batch size is 16, formed by two crops on each of eight processes. We use probe seed 0 and data seed 100 unless stated otherwise. Weight decay is $10^{-3}$ for the spatial readouts and $10^{-4}$ for the program readout. Training minimizes cross-entropy over non-void pixels. We evaluate the complete validation set every 5k updates to monitor training and report the final checkpoint after 40k updates.

At validation time, we resize the shorter image side to 256 and apply no stochastic augmentation. We evaluate $256\times256$ windows with stride 171. The logits from each window are resized bilinearly and averaged in overlapping regions. The merged logits are then resized to the original label resolution. All non-void pixels are evaluated, without flip-based test-time augmentation.

As a reference evaluation of the probe implementation, we also evaluate the frozen teachers at $512\times512$ with stride 341, following the DINOv3 linear protocol (Table~\ref{tab:teacher-reference}). These numbers concern teachers only and are not comparable with the matched $256\times256$ results.

\begin{table}[t]
\centering
\small
\caption{Reference evaluation of the probe implementation on frozen DINOv3 teachers at $512{\times}512$ (sliding window, stride 341), following the DINOv3 linear segmentation protocol~\citep{simeoni2025dinov3}. On ADE20K, our implementation matches the result of the official evaluation code for DINO-L (54.9 mIoU, the published value). These numbers validate the evaluation pipeline only: they concern teachers, not \mname, and are not comparable with the matched $256{\times}256$ results. Segmentation mIoU (\%, $\uparrow$).}
\label{tab:teacher-reference}
\setlength{\tabcolsep}{4pt}
\renewcommand{\arraystretch}{1.1}
\begin{tabular}{llrrr}
\toprule
Dataset & Evaluation & DINO-S & DINO-B & DINO-L \\
\midrule
ADE20K & DINOv3 evaluation code & --- & --- & 54.9 \\
ADE20K & our implementation of the same protocol & --- & --- & 55.1 \\
VOC & our implementation, DINOv3-style linear & 84.3 & 88.2 & 88.2 \\
Citysc. & our implementation, DINOv3-style linear & 67.3 & 71.6 & 73.5 \\
\bottomrule
\end{tabular}
\end{table}

\textbf{Multi-label classification.}
The multi-label experiment retains its original task-specific protocol. We train on the first 1000 images of the official training split and evaluate on the complete validation split. No online augmentation is applied. Representations from $Z_e$ and $Z_{\mathrm{out}}$ are mean-pooled over their spatial dimensions. The active prefix of $Z_q$ is aggregated using attention pooling.

We train for 20 epochs using AdamW with a learning rate of $10^{-3}$, weight decay $10^{-4}$, batch size 256, and seed 0. The objective is binary cross-entropy. We report mean average precision from the final epoch.

\textbf{Monocular depth estimation.}
For NYUv2, the dense readout predicts logits over 256 linearly spaced depth bins between $0.001\,\mathrm{m}$ and $10\,\mathrm{m}$. Given logit $\ell_b$ and bin centre $d_b$, we compute
\begin{equation}
    w_b
    =
    \frac{
        \max(\ell_b,0)+0.1
    }{
        \sum_{b'}
        \left(
            \max(\ell_{b'},0)+0.1
        \right)
    }.
\end{equation}
The continuous depth prediction is
\begin{equation}
    \hat{d}
    =
    \operatorname{clip}
    \left(
        \sum_b w_b d_b,\,
        0.001,\,
        10
    \right).
\end{equation}

We train the depth readouts for 38.4k updates, including 12.8k warmup updates, using AdamW with a learning rate of $3\times10^{-4}$, weight decay $10^{-4}$, a global batch size of 16, and seed 321. We report the final checkpoint. As a reference, the frozen teachers are additionally evaluated with the official DINOv3 NYUv2 depth protocol (Table~\ref{tab:depth-coco}, bottom block); all \mname comparisons use the matched protocol above.

\textbf{Matched external-tokenizer comparison.}
For the matched comparison with external tokenizers, all program representations use the same readout architecture and semantic-segmentation pipeline. Each readout is trained for 40k updates, including 1.5k warmup updates, with a global batch size of 16, a learning rate of $10^{-3}$, and weight decay $10^{-4}$. Budgets shown with filled markers in Fig.~\ref{fig:rate-quality-external} average three probe seeds; the remaining budgets use one. Depending on the input token dimension, the readouts contain 3.76--3.84M parameters and require 1.14--1.18 GFLOPs.

\textbf{Rate accounting.}
We report the mean active program length,
\begin{equation}
    \bar{L}
    =
    \frac{1}{N_{\mathrm{img}}}
    \sum_i L_i.
\end{equation}
For $H\times H$ inputs, we also estimate the idealized entropy rate from the empirical code distribution,
\begin{equation}
    \mathrm{bpp}
    =
    \frac{
        \sum_i\sum_{t=1}^{L_i}
        -\log_2 \pi_{c_{i,t}}
    }{
        N_{\mathrm{img}}H^2
    },
\end{equation}
where $c_{i,t}$ is the discrete code ID at position $t$ in the program for image $i$. This estimate includes the emitted token sequence and excludes the model, codebook, and sequence-length metadata.

\FloatBarrier
\section{Rate-quality curves}
\label{app:rate-quality}

\subsection{Internal rate-quality: adaptive vs.\ fixed length}
\label{app:rate-quality-internal}

\begin{figure}[t]
    \centering
    \includegraphics[width=\linewidth]{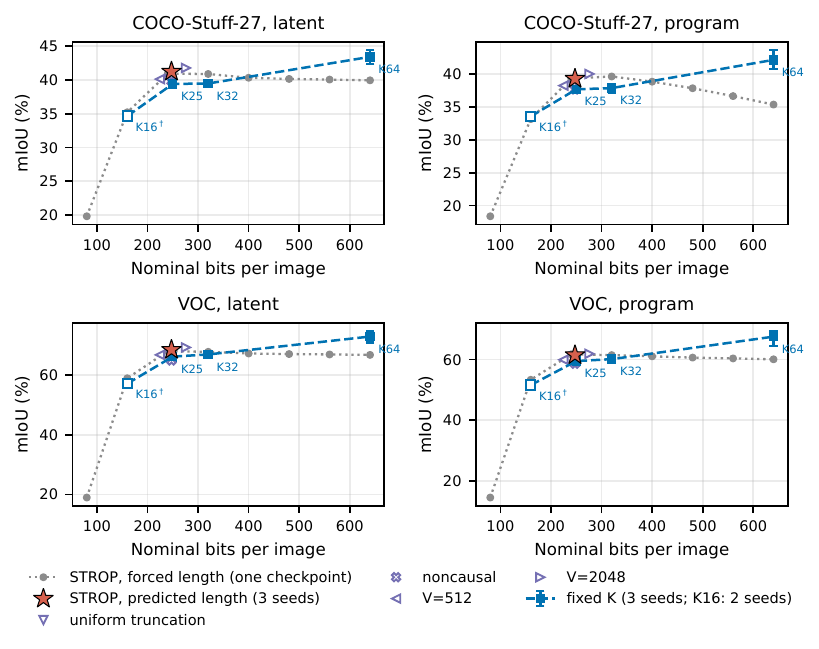}
    \caption{\textbf{Internal rate--quality.} Segmentation mIoU against nominal bits per $256{\times}256$ image for separately trained fixed-length models ($K{=}16$, 25, 32, 64; mean and range over 3 tokenizer seeds; $^\dagger$for $K{=}16$ over the 2 seeds that did not collapse), the adaptive \mname at its predicted length (star; mean and range over 3 tokenizer seeds), the same adaptive checkpoint truncated to forced lengths 8--64 (one checkpoint, final probes), and single ablations at their measured rate (open markers). Probes as in Table~\ref{tab:T1}.}
    \label{fig:rate-quality-coco}
\end{figure}

\textbf{Nominal bits.} Codes are drawn from an $V{=}1024$ codebook, so
$b_{\mathrm{code}}=\log_2 1024 = 10$ bits/token. The adaptive model also
carries a length field, $\lceil\log_2 64\rceil = 6$ bits, so a crop of length
$L$ costs $R(L) = 10L+6$ bits; the plotted mean rate uses $\mathbb{E}[L]$
($24.38$ on both COCO-Stuff-27 and VOC, $\bar{R} \approx 250$ bits). Fixed-$K$
baselines carry no length field: $R_K = 10K$ ($160$/$250$/$320$/$640$ bits for
$K{=}16/25/32/64$). These are bits per crop, not per source image -- larger
images yield multiple crops.

\textbf{Takeaway.} Figure~\ref{fig:rate-quality-coco} shows the effect directly. At matched rate, the learned length head matches or beats
both a uniform-allocation control and the fixed-$K$ frontier, consistent with
Table~\ref{tab:T1}. The adaptive curve saturates past its
elbow, while the fixed-$K$ frontier keeps climbing: forcing extra tokens
through a checkpoint trained for adaptive allocation stops helping, but
budgeting for more tokens from the start keeps buying quality. 

\subsection{External semantic rate-transfer comparison}
\label{app:rate-quality-external}

\begin{figure}[t]
    \centering
    \includegraphics[width=\linewidth]{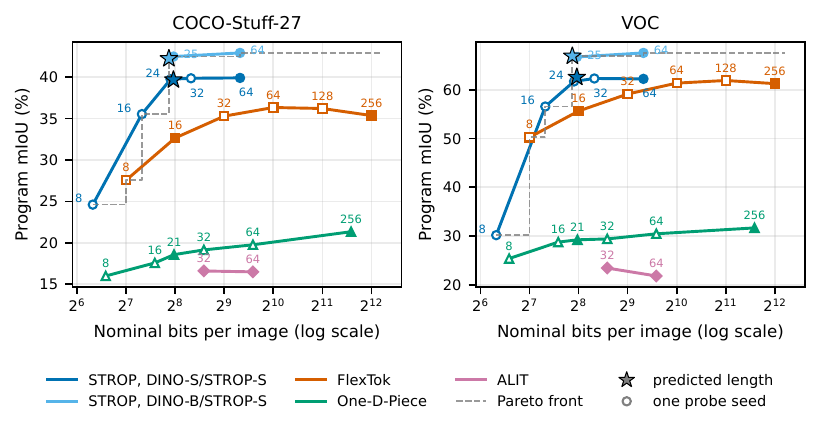}
    \caption{\textbf{Semantic rate--transfer frontier.} Program mIoU against the nominal number of code-index bits per $256{\times}256$ image for \mname and public adaptive tokenizers, under the same frozen-token probe (Table~\ref{tab:T2}). Each line connects one tokenizer's token budgets (labelled by token count); stars mark \mname's predicted length. Filled markers average three probe seeds, hollow markers use one. The dashed line is the Pareto front over all points.}
    \label{fig:rate-quality-external}
\end{figure}

Figure~\ref{fig:rate-quality-external} plots the resulting frontier for all
tokenizers. This is a semantic rate-transfer comparison, not a reconstruction
rate-distortion curve, and the rates are code-index accounting, not actual
compressed file size. The single matched-rate point closest to $\approx 250$
bits is reported in Table~\ref{tab:T2}; these curves show the
full frontier each tokenizer traces, and the caveat given there -- that this
is a system-level comparison across different architectures, training data,
and objectives, not a controlled ablation -- applies to the whole curve, not
just the matched point.

\FloatBarrier
\section{Ablation details}
\label{app:ablation-details}
Table~\ref{tab:ablation-absolute} reports the absolute scores behind the changes in Table~\ref{tab:T3}, with the spread over tokenizer seeds and higher-rate spatial references.

\begin{table}[h]
\centering
\caption{Absolute segmentation mIoU (\%, $\uparrow$) of every separately trained system in the ablation study (DINO-S / \msmall, COCO training), complementing the changes shown in Table~\ref{tab:T3}. Values are means over 3 tokenizer seeds, each averaged over 3 probe seeds, with the protocol of Table~\ref{tab:T1}; Mean averages the four benchmarks and its subscript is the standard deviation over tokenizer seeds. Bits are nominal index bits per image on COCO val; --- marks rates that were not measured (the continuous grid has no discrete rate). The high-rate spatial references encode the DINO patch grid with a fixed $8{\times}8$ or $16{\times}16$ token grid and are shown for context only. $^\dagger$One of the three seeds collapses: mean latent mIoU 3.0 / 32.5 / 32.8 for MSE only and 34.5 / 2.5 / 34.5 for $K{=}16$ across the three seeds. Stuff-27 denotes COCO-Stuff-27.}
\label{tab:ablation-absolute}
\setlength{\tabcolsep}{0.8pt}
\renewcommand{\arraystretch}{1.08}
\footnotesize
\begin{tabular*}{\linewidth}{@{\extracolsep{\fill}}lr rrrrr rrrrr@{}}
\toprule
 & & \multicolumn{5}{c}{Latent} & \multicolumn{5}{c}{Program} \\
\cmidrule(lr){3-7}\cmidrule(lr){8-12}
Variant & Bits & VOC & Stuff-27 & ADE20K & Citysc. & Mean & VOC & Stuff-27 & ADE20K & Citysc. & Mean \\
\midrule
\multicolumn{12}{l}{\emph{Teacher (frozen DINO)}} \\
DINO-S teacher & --- & 80.3 & 46.2 & 42.1 & 56.7 & 56.3 & --- & --- & --- & --- & --- \\
\midrule
\multicolumn{12}{l}{\emph{\mname (ours)}} \\
\mname (adaptive) & ${\approx}$250 & 68.6 & 41.2 & 28.0 & 36.8 & 43.7{\scriptsize$\pm$0.4} & 61.5 & 39.3 & 25.3 & 37.8 & 41.0{\scriptsize$\pm$0.4} \\
\midrule
\multicolumn{12}{l}{\emph{Fixed length}} \\
$K{=}16$$^\dagger$ & 160 & 39.3 & 23.3 & 13.3 & 19.5 & 23.8{\scriptsize$\pm$18.5} & 35.6 & 23.1 & 13.1 & 21.6 & 23.4{\scriptsize$\pm$17.5} \\
$K{=}25$ & 250 & 66.3 & 39.4 & 25.0 & 33.9 & 41.1{\scriptsize$\pm$1.5} & 59.5 & 37.7 & 23.4 & 35.7 & 39.1{\scriptsize$\pm$1.4} \\
$K{=}32$ & 320 & 67.0 & 39.5 & 25.8 & 34.9 & 41.8{\scriptsize$\pm$0.4} & 60.1 & 37.9 & 23.7 & 35.9 & 39.4{\scriptsize$\pm$0.4} \\
$K{=}64$ & 640 & 73.0 & 43.4 & 32.3 & 42.0 & 47.7{\scriptsize$\pm$1.8} & 67.6 & 42.2 & 29.5 & 42.9 & 45.6{\scriptsize$\pm$2.0} \\
\midrule
\multicolumn{12}{l}{\emph{Objective}} \\
+ pixel gradients & --- & 58.3 & 36.0 & 19.3 & 29.4 & 35.7{\scriptsize$\pm$0.4} & 49.8 & 32.1 & 17.2 & 30.6 & 32.4{\scriptsize$\pm$0.3} \\
cosine loss only & --- & 65.6 & 39.6 & 25.4 & 34.2 & 41.2{\scriptsize$\pm$2.3} & 59.0 & 37.6 & 23.4 & 35.5 & 38.9{\scriptsize$\pm$2.1} \\
MSE loss only$^\dagger$ & --- & 38.0 & 23.1 & 12.0 & 18.1 & 22.8{\scriptsize$\pm$17.1} & 34.1 & 22.1 & 11.8 & 20.2 & 22.1{\scriptsize$\pm$16.5} \\
\midrule
\multicolumn{12}{l}{\emph{Ordering}} \\
noncausal generator & ${\approx}$248 & 65.0 & 39.8 & 25.5 & 33.4 & 40.9{\scriptsize$\pm$2.9} & 58.6 & 37.7 & 23.4 & 34.3 & 38.5{\scriptsize$\pm$2.8} \\
spatial $5{\times}5$ grid & 250 & 68.3 & 41.2 & 27.9 & 36.8 & 43.6{\scriptsize$\pm$0.5} & 61.5 & 39.0 & 25.8 & 38.7 & 41.3{\scriptsize$\pm$0.3} \\
\midrule
\multicolumn{12}{l}{\emph{Curriculum}} \\
joint from start & --- & 63.9 & 38.2 & 23.9 & 31.7 & 39.4{\scriptsize$\pm$1.6} & 57.7 & 35.7 & 22.2 & 33.3 & 37.2{\scriptsize$\pm$1.1} \\
uniform truncation & ${\approx}$244 & 67.3 & 40.2 & 26.5 & 35.1 & 42.2{\scriptsize$\pm$0.8} & 60.9 & 38.3 & 24.5 & 36.0 & 39.9{\scriptsize$\pm$0.5} \\
abrupt handoff & --- & 68.7 & 41.2 & 27.9 & 36.3 & 43.5{\scriptsize$\pm$0.1} & 61.4 & 39.3 & 25.6 & 37.7 & 41.0{\scriptsize$\pm$0.3} \\
no oracle warm-up & --- & 68.6 & 41.4 & 28.1 & 36.8 & 43.7{\scriptsize$\pm$0.2} & 61.2 & 39.4 & 26.0 & 37.3 & 41.0{\scriptsize$\pm$0.1} \\
\midrule
\multicolumn{12}{l}{\emph{Vocabulary}} \\
$V{=}512$ & ${\approx}$225 & 66.8 & 40.1 & 26.5 & 34.6 & 42.0{\scriptsize$\pm$0.3} & 60.0 & 38.3 & 24.0 & 36.0 & 39.6{\scriptsize$\pm$0.3} \\
$V{=}2048$ & ${\approx}$276 & 69.3 & 41.8 & 29.1 & 38.6 & 44.7{\scriptsize$\pm$0.2} & 61.9 & 40.0 & 26.5 & 39.3 & 41.9{\scriptsize$\pm$0.5} \\
\midrule
\multicolumn{12}{l}{\emph{High-rate spatial references}} \\
grid $8{\times}8$, VQ & 640 & 75.0 & 44.3 & 34.2 & 45.9 & 49.8{\scriptsize$\pm$0.6} & 68.1 & 43.0 & 31.5 & 46.2 & 47.2{\scriptsize$\pm$1.1} \\
grid $8{\times}8$, VQ, conv. & 640 & 71.0 & 42.2 & 30.8 & 42.8 & 46.7{\scriptsize$\pm$0.1} & 68.1 & 43.3 & 30.6 & 46.5 & 47.1{\scriptsize$\pm$0.0} \\
grid $16{\times}16$, VQ & 2560 & 78.0 & 45.5 & 38.4 & 51.6 & 53.4{\scriptsize$\pm$0.1} & 72.2 & 44.9 & 35.0 & 52.9 & 51.2{\scriptsize$\pm$0.8} \\
grid $8{\times}8$, continuous & --- & 78.3 & 45.8 & 39.2 & 52.6 & 54.0{\scriptsize$\pm$0.1} & 75.6 & 47.6 & 37.6 & 54.5 & 53.8{\scriptsize$\pm$0.2} \\
\bottomrule
\end{tabular*}
\end{table}

\FloatBarrier
\section{Replication across configurations}
\label{app:replication}
Table~\ref{tab:cross-model} repeats the main contrasts with a different training set (ImageNet), a larger teacher (DINO-B) and a larger generator (\mmedium). Each configuration is trained with three tokenizer seeds.

\begin{table}[t]
\centering
\caption{Replication of the main contrasts across training data, teacher size and generator size. Each column pair is one configuration (training data, encoder / model); cells give the change in COCO-Stuff-27 segmentation mIoU (points) for the latent (L) and program (P) probes, shaded as in Table~\ref{tab:T3} (red: worse, blue: better). Means over 3 tokenizer seeds; a superscript gives the number of seeds that agree in sign with the mean when fewer than all three do; --- marks contrasts that were not run. The first configuration uses the protocol of Table~\ref{tab:T1}; the others use one probe seed. Bits give the nominal rate of the adaptive and the rate-matched fixed tokenizer on COCO val (the ImageNet-trained length was matched on ImageNet val, so on COCO its adaptive rate is higher). Order: teacher cosine of the interpreted latents before $\rightarrow$ after shuffling the program at test time (mean over tokenizer seeds; 5,000 COCO val images). ``Adaptive $-$ fixed'' compares with the separately trained fixed-length tokenizer matched to the adaptive rate; ``sequential $-$ grid'' compares a sequential and a $5{\times}5$ spatial tokenizer, both with 25 tokens.}
\label{tab:cross-model}
\definecolor{abneg}{RGB}{178,24,43}\definecolor{abpos}{RGB}{33,102,172}
\setlength{\tabcolsep}{2.5pt}
\renewcommand{\arraystretch}{1.12}
\footnotesize
\begin{tabular*}{\linewidth}{@{\extracolsep{\fill}}l rr rr rr rr@{}}
\toprule
 & \multicolumn{2}{c}{COCO} & \multicolumn{2}{c}{ImageNet} & \multicolumn{2}{c}{COCO} & \multicolumn{2}{c}{COCO} \\
 & \multicolumn{2}{c}{DINO-S/\msmall} & \multicolumn{2}{c}{DINO-S/\msmall} & \multicolumn{2}{c}{DINO-B/\msmall} & \multicolumn{2}{c}{DINO-S/\mmedium} \\
\cmidrule(lr){2-3}\cmidrule(lr){4-5}\cmidrule(lr){6-7}\cmidrule(lr){8-9}
Contrast & L & P & L & P & L & P & L & P \\
\midrule
adaptive $-$ fixed & \cellcolor{abpos!21}$+$1.8 & \cellcolor{abpos!20}$+$1.6 & \cellcolor{abpos!25}$+$2.4 & \cellcolor{abpos!23}$+$2.1 & \cellcolor{abpos!35}$+$4.8 & \cellcolor{abpos!31}$+$3.8 & --- & --- \\
+ pixel gradients & \cellcolor{abneg!36}$-$5.3 & \cellcolor{abneg!42}$-$7.2 & \cellcolor{abneg!34}$-$4.6 & \cellcolor{abneg!39}$-$6.0 & --- & --- & --- & --- \\
sequential $-$ grid, $K{=}25$ & \cellcolor{abneg!21}$-$1.8 & \cellcolor{abneg!18}$-$1.3 & --- & --- & \cellcolor{abneg!28}$-$3.2 & \cellcolor{abneg!25}$-$2.5 & --- & --- \\
noncausal generator & \cellcolor{abneg!19}$-$1.4$^{2/3}$ & \cellcolor{abneg!20}$-$1.6$^{2/3}$ & --- & --- & --- & --- & \cellcolor{abneg!12}$-$0.5 & \cellcolor{abneg!24}$-$2.3 \\
joint from start & \cellcolor{abneg!28}$-$3.1 & \cellcolor{abneg!30}$-$3.6 & --- & --- & --- & --- & \cellcolor{abneg!15}$-$0.8$^{2/3}$ & \cellcolor{abneg!15}$-$0.9$^{2/3}$ \\
\midrule
Bits (adapt.\,/\,fixed) & \multicolumn{2}{c}{${\approx}$250 / 250} & \multicolumn{2}{c}{269--280 / 250} & \multicolumn{2}{c}{234--236 / 230} & \multicolumn{2}{c}{259--266 / ---} \\
Order (Cos) & \multicolumn{2}{c}{0.86 $\rightarrow$ 0.53} & \multicolumn{2}{c}{0.85 $\rightarrow$ 0.49} & \multicolumn{2}{c}{0.81 $\rightarrow$ 0.53} & \multicolumn{2}{c}{0.85 $\rightarrow$ 0.47} \\
\bottomrule
\end{tabular*}
\end{table}

\FloatBarrier

\section{Alignment}\label{app:alignment}
Table~\ref{tab:detailed-test-metrics} expands Table~\ref{tab:test-metrics} with per-dataset latent alignment and codebook statistics.

Table~\ref{tab:test-metrics} compares the interpreted feature field of each variant with frozen DINOv3 features. Under the same 64-token, 1024-entry bottleneck, DINO-S is easiest to compress: DINO-S/\mname-S gives the best COCO alignment ($.855$ cosine, $R^2=.734$), while DINO-S/\mname-M is best on ImageNet ($.851$, $R^2=.727$). DINO-B remains around $.80$ cosine and $R^2\approx.64$, and DINO-L drops to $\sim.72/.53$, consistent with a fixed-rate bottleneck facing higher-dimensional teacher features. Larger generators mainly increase vocabulary coverage: \mname-M uses 91--98\% of the codebook, but this does not monotonically improve alignment, e.g., COCO DINO-S/\mname-S outperforms DINO-S/\mname-M despite lower codebook usage.

\textbf{Alignment and transfer.}
Table~\ref{tab:T5b} varies the teacher with the \msmall generator fixed. Teacher cosine is $0.858$ for DINO-S, $0.810$ for DINO-B, and $0.747$ for DINO-L, yet DINO-B segments better than DINO-S ($44.8$ against $41.5$ latent mIoU on COCO-Stuff-27) and DINO-L matches DINO-B. The DINO-L teacher itself is only slightly better than DINO-B on these benchmarks. Across the full model grid, the larger \mmedium generator does not consistently help under the shared training schedule (Appendix~\ref{app:downstream}).

\textbf{Compression.}
With $256{\times}256$ inputs and a patch size of $16$, every encoder emits a
fixed grid of $256$ patch tokens (with dimension $D$) per image, giving per-image feature tensors
of shape $256{\times}384$ (DINO-S), $256{\times}768$ (DINO-B), and $256{\times}1024$ (DINO-L).
\mname compresses each of these backbone-dependent tensors into an
adaptive sequence of at most $64$ discrete tokens drawn from a codebook of size
$V{=}1024$ ($d_c{=}16$), corresponding to a hard ceiling of
$64 \cdot \log_2 V = 640$ bits per image regardless of encoder choice. Against
the continuous float32 source representations of $256{\cdot}D{\cdot}32$ bits,
this yields worst-case (full $64$-token) compression ratios of
${\sim}4{,}915{\times}$ (DINO-S), ${\sim}9{,}830{\times}$ (DINO-B), and
${\sim}13{,}107{\times}$ (DINO-L), with the use of length head typically leading to shorter sequences and higher
ratios. The total source entropy scales linearly with $D$, but the bit budget is held constant, so
larger backbones face a strictly harder rate--distortion problem: DINO-L must
compress $2.67{\times}$ more raw feature dimensions than DINO-S into the same
$640$-bit envelope. %

\begin{table}[t]
\centering
\small
\caption{Teacher size at fixed generator capacity. The five main variants (Table~\ref{tab:variants}) pair DINO-L only with \mmedium, which confounds teacher and generator size; here the teacher varies with \msmall fixed (COCO training). Bits is the nominal rate and Cos the cosine similarity between the interpreted latents and the frozen teacher's patch features, both on 5,000 COCO val images. Segmentation mIoU (\%, $\uparrow$) is reported for the frozen teacher (an upper bound) and for the \mname latent and program readouts, with probes as in Table~\ref{tab:T1}; one checkpoint and one probe seed per row. A second DINO-L tokenizer seed gives 44.8 (latent) and 42.6 (program) on COCO-Stuff-27 at 232.8 bits. Stuff-27 denotes COCO-Stuff-27. The best non-teacher value in each column is \textbf{bolded}.}
\label{tab:T5b}
\setlength{\tabcolsep}{4pt}
\renewcommand{\arraystretch}{1.1}
\begin{tabular}{lrr rrr rrr}
\toprule
 & & & \multicolumn{3}{c}{Stuff-27 mIoU $\uparrow$} & \multicolumn{3}{c}{VOC mIoU $\uparrow$} \\
\cmidrule(lr){4-6}\cmidrule(lr){7-9}
Encoder & Bits & Cos$\uparrow$ & Teacher & Latent & Program & Teacher & Latent & Program \\
\midrule
DINO-S & ${\approx}$250 & \textbf{0.858} & 46.1 & 41.5 & 39.7 & 80.2 & 68.8 & 62.5 \\
DINO-B & ${\approx}$234 & 0.810 & 51.6 & \textbf{44.8} & 42.3 & 84.3 & \textbf{72.9} & 66.6 \\
DINO-L & ${\approx}$234 & 0.747 & 52.7 & \textbf{44.8} & \textbf{42.8} & 84.7 & \textbf{72.9} & \textbf{67.7} \\
\bottomrule
\end{tabular}
\end{table}

\begin{table}[t]
\centering
\caption{Latent reconstruction quality and codebook utilization, averaged over five evaluation datasets (VOC, COCO-Stuff-27, ADE20K, Cityscapes, NYUv2). Cos and $R^2$ are measured against frozen DINO teacher patches; RMSE is computed per patch-feature scalar. V\% is the fraction of the 1024-entry codebook used at least once by active program tokens; Eff.\% is the entropy-effective utilization (Appendix~\ref{app:evaluation_protocol}).}
\label{tab:test-metrics}
\setlength{\tabcolsep}{3pt}
\renewcommand{\arraystretch}{1.1}
\footnotesize
\begin{tabular*}{\linewidth}{@{\extracolsep{\fill}}l rrrrr rrrrr@{}}
\toprule
         & \multicolumn{5}{c}{Trained on COCO} & \multicolumn{5}{c}{Trained on ImageNet} \\
\cmidrule(lr){2-6}\cmidrule(lr){7-11}
Variant  & Cos$\uparrow$ & $R^2\uparrow$ & RMSE$\downarrow$ & V\%$\uparrow$ & Eff.\%$\uparrow$ & Cos$\uparrow$ & $R^2\uparrow$ & RMSE$\downarrow$ & V\%$\uparrow$ & Eff.\%$\uparrow$ \\
\midrule
DINO-S/\msmall & 0.855 & 0.734 & 0.196 & 56.1 & 44.7 & 0.849 & 0.724 & 0.199 & 56.2 & 44.5 \\
DINO-S/\mmedium & 0.835 & 0.697 & 0.208 & 98.0 & 78.6 & 0.851 & 0.727 & 0.198 & 98.1 & 72.0 \\
DINO-B/\msmall & 0.801 & 0.643 & 0.246 & 53.2 & 41.2 & 0.763 & 0.583 & 0.266 & 77.2 & 60.0 \\
DINO-B/\mmedium & 0.801 & 0.642 & 0.246 & 94.8 & 78.7 & 0.801 & 0.643 & 0.246 & 91.2 & 75.3 \\
DINO-L/\mmedium & 0.725 & 0.526 & 0.198 & 95.9 & 81.7 & 0.723 & 0.527 & 0.198 & 92.2 & 72.9 \\
\bottomrule
\end{tabular*}
\end{table}

\begin{table}[t]
\centering
\caption{Latent alignment per evaluation dataset. Cosine similarity and $R^2$ between the interpreter's predicted patch latents and frozen DINO teacher patch features on the validation split of each evaluation dataset. Codebook statistics are reported in Table~\ref{tab:test-metrics}. Stuff-27 denotes COCO-Stuff-27.}
\label{tab:detailed-test-metrics}
\setlength{\tabcolsep}{2.5pt}
\renewcommand{\arraystretch}{1.1}
\footnotesize
\begin{tabular*}{\linewidth}{@{\extracolsep{\fill}}l rrrrr rrrrr@{}}
\toprule
 & \multicolumn{5}{c}{Cos$\uparrow$} & \multicolumn{5}{c}{$R^2\uparrow$} \\
\cmidrule(lr){2-6}\cmidrule(lr){7-11}
Variant & VOC & Stuff-27 & ADE20K & Citysc. & NYUv2 & VOC & Stuff-27 & ADE20K & Citysc. & NYUv2 \\
\midrule
\multicolumn{11}{l}{\emph{Trained on COCO}} \\
DINO-S/\msmall & 0.859 & 0.858 & 0.854 & 0.849 & 0.853 & 0.742 & 0.742 & 0.731 & 0.722 & 0.731 \\
DINO-S/\mmedium & 0.841 & 0.839 & 0.835 & 0.834 & 0.825 & 0.709 & 0.707 & 0.696 & 0.693 & 0.679 \\
DINO-B/\msmall & 0.804 & 0.810 & 0.792 & 0.799 & 0.798 & 0.650 & 0.661 & 0.629 & 0.639 & 0.637 \\
DINO-B/\mmedium & 0.808 & 0.810 & 0.794 & 0.797 & 0.794 & 0.656 & 0.661 & 0.631 & 0.633 & 0.630 \\
DINO-L/\mmedium & 0.745 & 0.748 & 0.714 & 0.713 & 0.703 & 0.557 & 0.565 & 0.510 & 0.505 & 0.491 \\
\midrule
\multicolumn{11}{l}{\emph{Trained on ImageNet}} \\
DINO-S/\msmall & 0.863 & 0.854 & 0.854 & 0.828 & 0.848 & 0.750 & 0.734 & 0.731 & 0.685 & 0.721 \\
DINO-S/\mmedium & 0.865 & 0.854 & 0.857 & 0.834 & 0.844 & 0.753 & 0.734 & 0.736 & 0.694 & 0.716 \\
DINO-B/\msmall & 0.782 & 0.768 & 0.761 & 0.744 & 0.759 & 0.614 & 0.594 & 0.581 & 0.551 & 0.576 \\
DINO-B/\mmedium & 0.819 & 0.805 & 0.803 & 0.780 & 0.796 & 0.674 & 0.654 & 0.647 & 0.605 & 0.636 \\
DINO-L/\mmedium & 0.759 & 0.739 & 0.723 & 0.688 & 0.708 & 0.582 & 0.554 & 0.528 & 0.467 & 0.504 \\
\bottomrule
\end{tabular*}
\end{table}

\FloatBarrier
\section{Pairwise Token Synergies Align with Semantic Regions}
\label{app:pairwise-token-synergies}

\begin{figure}[h]
    \centering
    \includegraphics[width=1.0\linewidth]{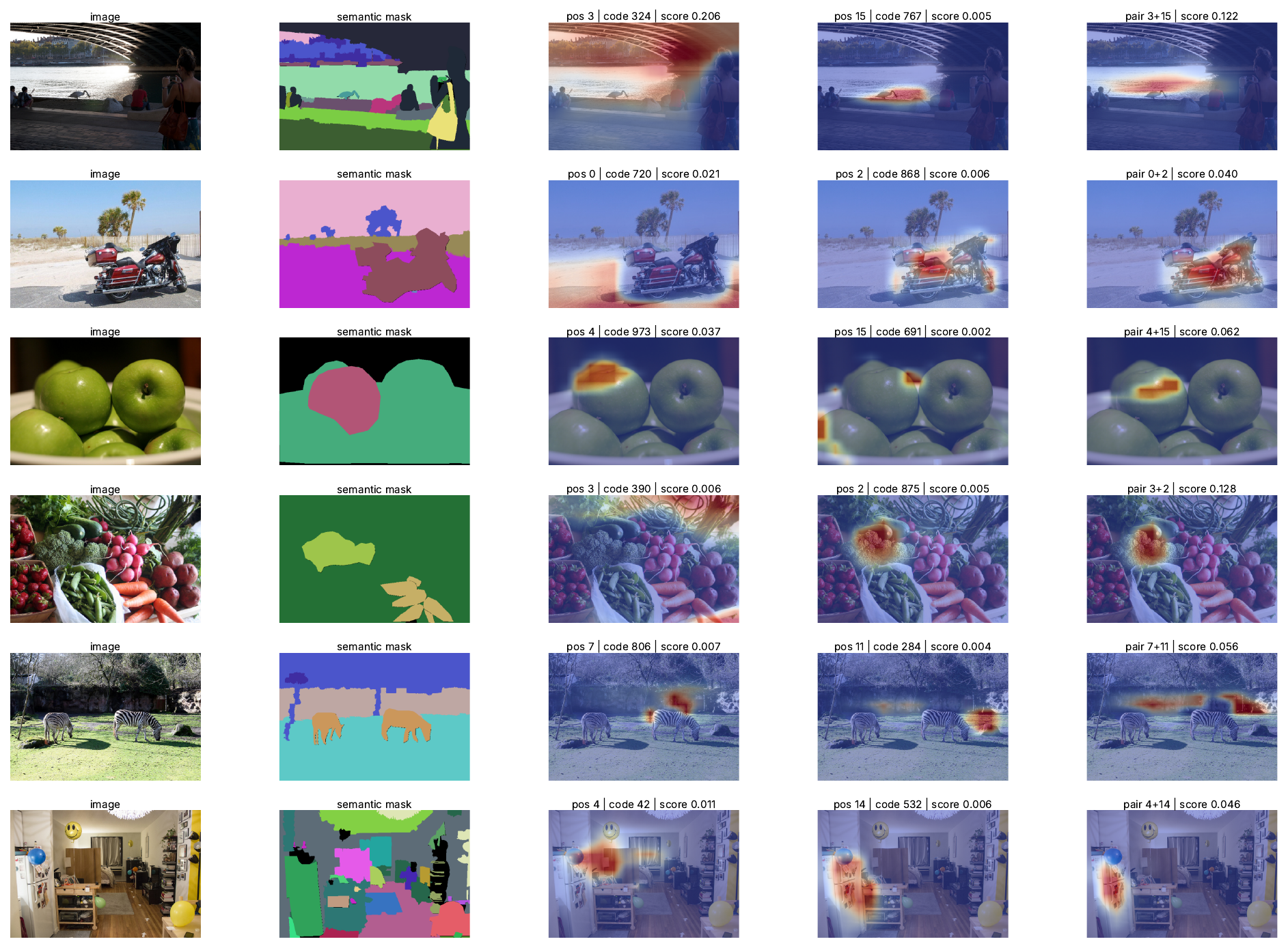}
\caption{\textbf{Pairwise program-token synergies.}
      We erase pairs of discrete program tokens and reinterpret the remaining sequence, then visualize the resulting change in the DINO-aligned patch field. The selected COCO-Stuff-27 examples show that high-synergy token pairs often produce localized responses aligned
  with semantic mask regions.
  Heatmaps are normalized per panel for visualization; pair scores are raw attribution magnitudes.}
      \label{fig:pair-synergy-coco-stuff}
\end{figure}

Pairwise interactions provide another view of the structure learned by the discrete image programs. As shown in Fig.~\ref{fig:pair-synergy-coco-stuff}, erasing two tokens jointly can produce localized changes that align with coherent semantic regions.

\FloatBarrier
\section{Downstream probe details}
\label{app:downstream}
We report downstream probes of the teacher, interpreted-field, and program representations for all model variants: Tables~\ref{tab:multilabel-coco-full}--\ref{tab:multilabel-imagenet-full} give multi-label classification under its original protocol, Table~\ref{tab:multiprobe-coco-full} gives semantic segmentation, and Table~\ref{tab:depth-coco} gives NYUv2 depth (protocols in Appendix~\ref{app:evaluation_protocol}).
The COCO-trained DINO-B/\msmall model segments best, with $47.3$ mean mIoU for the latent readout against $61.3$ for its frozen teacher. 
For NYUv2 depth, the program readouts of DINO-L/\mmedium and DINO-B/\msmall give the lowest errors among \mname{} variants, still above those of the frozen teachers (Table~\ref{tab:depth-coco}).

\textbf{Program and interpreted field.}
The program readout is a small transformer over the active codes. On segmentation it trails the latent readout by $7$ points on VOC and by $2$--$3$ points on COCO-Stuff-27 and ADE20K, and it is ahead on Cityscapes (Table~\ref{tab:T1}). On depth it reaches a slightly lower RMSE than the latent readout for all COCO-trained variants. The program falls far behind only in the multi-label experiment, which keeps its original attention-pooled linear readout. Most of the dense-prediction information is therefore still present in the discrete program.

\begin{figure}[t]
     \centering
     \begin{subfigure}[b]{0.49\textwidth}
         \centering
         \includegraphics[width=\textwidth]{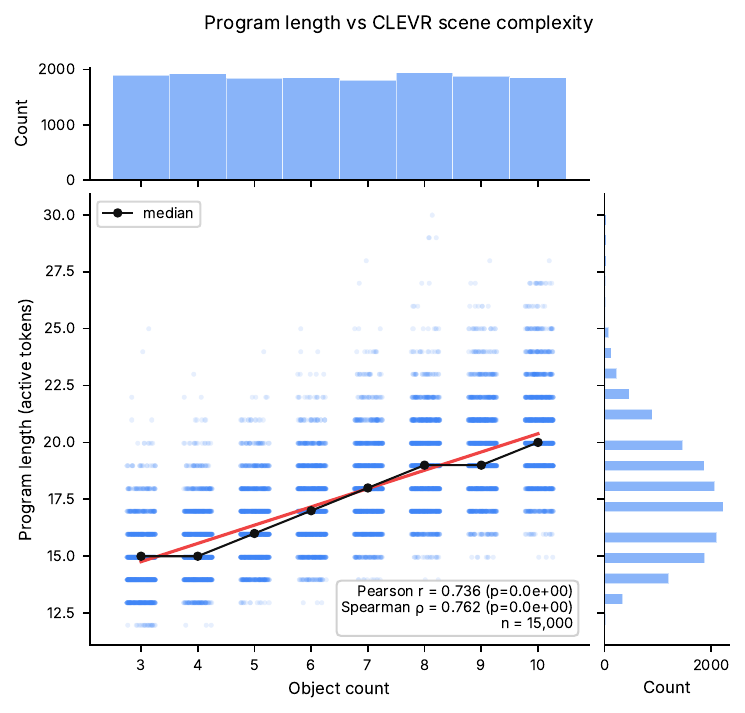}
         \caption{CLEVR; codebook size 32.}
         \label{fig:clevr_complexity}
     \end{subfigure}
     \hfill
     \begin{subfigure}[b]{0.49\textwidth}
         \centering
         \includegraphics[width=\textwidth]{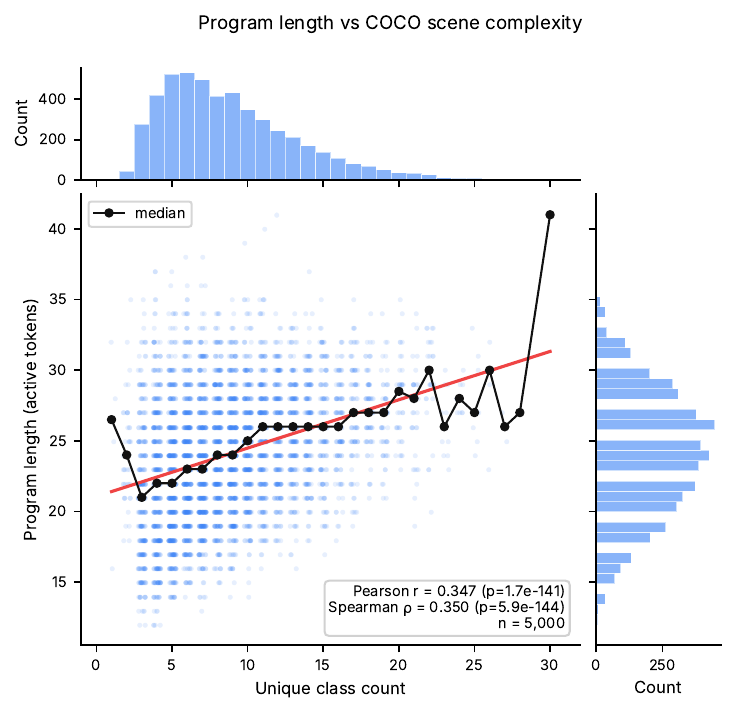}
         \caption{COCO; codebook size 1024.}
         \label{fig:coc_object_complexity}
     \end{subfigure}
        \caption{Program length and scene complexity on CLEVR and COCO. (a) On CLEVR, \mname produces longer programs as the number of objects increases. (b) On COCO, length also grows with the number of annotated classes, but the relation is weaker and noisier (Pearson $r = 0.35$); medians at the highest class counts rest on few images. Class counts are taken from the full-image annotations, while the model sees a $256{\times}256$ view.}
        \label{fig:clevr_main}
\end{figure}

\textbf{Program length tracks scene complexity.}
\mname{} allocates longer programs to richer scenes. On CLEVR the
correlation between active program length and ground-truth object count is
strong (Pearson $r \approx 0.74$; Figure~\ref{fig:clevr_complexity}). On
COCO, where scene complexity is a noisier quantity, the correlation against
the number of unique semantic classes is positive but weaker
(Figure~\ref{fig:coc_object_complexity}). The same trend holds against the
spatial variance of patch-level DINO features
(Figure~\ref{fig:dino_complexity}).
CLEVR ablations are in Appendix~\ref{app:clevr}.

\input{body/figure-segmentation-probes-coco}

\begin{table}[t]
\centering
\caption{Linear probe mAP (\%, $\uparrow$) on multi-label classification.
\mname trained on COCO; Stuff-27 = COCO-Stuff-27;
Mean = mean of Stuff-27 and VOC mAP.
Teacher uses mean-pooled DINO patch features;
latent is mean-pooled over the interpreter's spatial latent field;
program is attention-pooled over the active program token vectors $z_q$.
Best non-teacher value per column is \textbf{bolded}.
Probes follow the original multi-label protocol (1{,}000 training images; Appendix~\ref{app:evaluation_protocol}).}
\label{tab:multilabel-coco-full}
\setlength{\tabcolsep}{4pt}
\renewcommand{\arraystretch}{1.1}
\begin{tabular*}{\linewidth}{@{\extracolsep{\fill}}lllrrr@{}}
\toprule
 & & & \multicolumn{3}{c}{Multi-label mAP $\uparrow$} \\
\cmidrule(lr){4-6}
Encoder & Model & Repr.   & Stuff-27 & VOC  & Mean \\
\midrule
\emph{Teacher (frozen DINO)} & & & & & \\
DINO-S & ---  & teacher   & 56.2 & 59.4 & 57.8 \\
DINO-B & ---  & teacher   & 61.2 & 74.7 & 67.9 \\
DINO-L & ---  & teacher   & 59.7 & 68.4 & 64.0 \\
\midrule
\emph{\mname (ours)} & & & & & \\
DINO-S & \msmall   & latent    & 56.6 & 61.5 & 59.1 \\
DINO-S & \msmall   & program   & 34.9 & 17.8 & 26.3 \\
DINO-S & \mmedium  & latent    & 56.5 & 59.4 & 58.0 \\
DINO-S & \mmedium  & program   & 30.5 & 13.3 & 21.9 \\
\cmidrule(l){1-6}
DINO-B & \msmall   & latent    & \textbf{61.7} & 74.1 & 67.9 \\
DINO-B & \msmall   & program   & 36.1 & 18.5 & 27.3 \\
DINO-B & \mmedium  & latent    & 61.6 & \textbf{75.2} & \textbf{68.4} \\
DINO-B & \mmedium  & program   & 31.3 & 14.7 & 23.0 \\
\cmidrule(l){1-6}
DINO-L & \mmedium  & latent    & 59.9 & 68.3 & 64.1 \\
DINO-L & \mmedium  & program   & 34.5 & 16.1 & 25.3 \\
\bottomrule
\end{tabular*}
\end{table}

\begin{table}[t]
\centering
\caption{Linear probe mAP (\%, $\uparrow$) on multi-label classification.
\mname trained on ImageNet; Stuff-27 = COCO-Stuff-27;
Mean = mean of Stuff-27 and VOC mAP.
Teacher uses mean-pooled DINO patch features;
latent is mean-pooled over the interpreter's spatial latent field;
program is attention-pooled over the active program token vectors $z_q$.
Selected variants; for all results please refer to Appendix Table \ref{tab:multilabel-imagenet-full}.
Best non-teacher value per column is \textbf{bolded}.
Probes follow the original multi-label protocol (1{,}000 training images; Appendix~\ref{app:evaluation_protocol}).}
\label{tab:multilabel-imagenet-compact}
\setlength{\tabcolsep}{4pt}
\renewcommand{\arraystretch}{1.1}
\begin{tabular*}{\linewidth}{@{\extracolsep{\fill}}lllrrr@{}}
\toprule
Encoder & Model & Repr.   & Stuff-27 & VOC  & Mean \\
\midrule
\emph{Teacher (frozen DINO)} & & & & & \\
DINO-S & ---  & teacher   & 56.2 & 59.4 & 57.8 \\
DINO-B & ---  & teacher   & 61.2 & 74.7 & 67.9 \\
\midrule
\emph{\mname (ours)} & & & & & \\
DINO-S & \msmall   & latent    & 56.6 & 61.5 & 59.1 \\
DINO-S & \msmall   & program   & 34.9 & 17.8 & 26.3 \\
\cmidrule(l){1-6}
DINO-B & \msmall   & latent    & \textbf{61.7} & 74.1 & 67.9 \\
DINO-B & \msmall   & program   & 36.1 & 18.5 & 27.3 \\
DINO-B & \mmedium  & latent    & 61.6 & \textbf{75.2} & \textbf{68.4} \\
DINO-B & \mmedium  & program   & 31.3 & 14.7 & 23.0 \\
\bottomrule
\end{tabular*}
\vspace{-5mm}
\end{table}

\begin{table}[t]
\centering
\caption{Linear probe mAP (\%, $\uparrow$) on multi-label classification.
\mname trained on ImageNet; Stuff-27 = COCO-Stuff-27;
Mean = mean of Stuff-27 and VOC mAP.
Teacher uses mean-pooled DINO patch features;
latent is mean-pooled over the interpreter's spatial latent field;
program is attention-pooled over the active program token vectors $z_q$.
Best non-teacher value per column is \textbf{bolded}.
Probes follow the original multi-label protocol (1{,}000 training images; Appendix~\ref{app:evaluation_protocol}).}
\label{tab:multilabel-imagenet-full}
\setlength{\tabcolsep}{4pt}
\renewcommand{\arraystretch}{1.1}
\begin{tabular*}{\linewidth}{@{\extracolsep{\fill}}lllrrr@{}}
\toprule
 & & & \multicolumn{3}{c}{Multi-label mAP $\uparrow$} \\
\cmidrule(lr){4-6}
Encoder & Model & Repr.   & Stuff-27 & VOC  & Mean \\
\midrule
\emph{Teacher (frozen DINO)} & & & & & \\
DINO-S & ---  & teacher   & 56.2 & 59.4 & 57.8 \\
DINO-B & ---  & teacher   & 61.2 & 74.7 & 67.9 \\
DINO-L & ---  & teacher   & 59.7 & 68.4 & 64.0 \\
\midrule
\emph{\mname (ours)} & & & & & \\
DINO-S & \mname-S  & latent    & 55.7 & 57.6 & 56.6 \\
DINO-S & \mname-S  & program   & 32.3 & 15.0 & 23.7 \\
DINO-S & \mname-M  & latent    & 56.0 & 59.7 & 57.9 \\
DINO-S & \mname-M  & program   & 34.1 & 14.3 & 24.2 \\
\cmidrule(l){1-6}
DINO-B & \mname-S  & latent    & 60.2 & 69.6 & 64.9 \\
DINO-B & \mname-S  & program   & 32.6 & 14.0 & 23.3 \\
DINO-B & \mname-M  & latent    & \textbf{60.5} & \textbf{73.4} & \textbf{67.0} \\
DINO-B & \mname-M  & program   & 30.1 & 12.6 & 21.3 \\
\cmidrule(l){1-6}
DINO-L & \mname-M  & latent    & 59.1 & 68.2 & 63.6 \\
DINO-L & \mname-M  & program   & 32.2 & 13.9 & 23.1 \\
\bottomrule
\end{tabular*}
\end{table}

\begin{table}[t]
\centering
\caption{Segmentation mIoU (\%, $\uparrow$) of frozen representations for all model variants, trained on COCO or ImageNet. All rows use the matched $256{\times}256$ protocol of Appendix~\ref{app:evaluation_protocol} (full training split, 40k updates, final checkpoint, probe seed 0). Teacher rows probe frozen DINO patch features as an upper bound and do not depend on the training set of \mname. Latent probes the interpreter's spatial latent field; program probes the active program token vectors $z_q$. Stuff-27 denotes COCO-Stuff-27. The best non-teacher value in each column is \textbf{bolded}.}
\label{tab:multiprobe-coco-full}
\setlength{\tabcolsep}{2.5pt}
\renewcommand{\arraystretch}{1.1}
\footnotesize
\begin{tabular*}{\linewidth}{@{\extracolsep{\fill}}lll rrrr rrrr@{}}
\toprule
 & & & \multicolumn{4}{c}{Trained on COCO} & \multicolumn{4}{c}{Trained on ImageNet} \\
\cmidrule(lr){4-7}\cmidrule(lr){8-11}
Encoder & Model & Repr. & VOC & Stuff-27 & ADE20K & Citysc. & VOC & Stuff-27 & ADE20K & Citysc. \\
\midrule
\multicolumn{11}{l}{\emph{Teacher (frozen DINO)}} \\
DINO-S & --- & teacher & 80.2 & 46.1 & 42.1 & 56.6 & 80.2 & 46.1 & 42.1 & 56.6 \\
DINO-B & --- & teacher & 84.3 & 51.6 & 47.1 & 62.1 & 84.3 & 51.6 & 47.1 & 62.1 \\
DINO-L & --- & teacher & 84.7 & 52.7 & 50.5 & 63.5 & 84.7 & 52.7 & 50.5 & 63.5 \\
\midrule
\multicolumn{11}{l}{\emph{\mname (ours)}} \\
DINO-S & \msmall & latent & 68.8 & 41.5 & 29.0 & 37.0 & 68.7 & 40.1 & 27.9 & 35.8 \\
DINO-S & \msmall & program & 62.5 & 39.7 & 25.3 & 38.8 & 64.1 & 39.4 & 25.1 & \textbf{37.9} \\
DINO-S & \mmedium & latent & 61.2 & 36.3 & 22.2 & 29.7 & 65.7 & 38.3 & 24.9 & 31.5 \\
DINO-S & \mmedium & program & 54.1 & 32.0 & 20.4 & 30.4 & 58.9 & 35.4 & 21.2 & 32.9 \\
\cmidrule(l){1-11}
DINO-B & \msmall & latent & \textbf{72.9} & \textbf{44.8} & \textbf{30.7} & 40.8 & 66.6 & 38.9 & 24.4 & 29.7 \\
DINO-B & \msmall & program & 66.6 & 42.3 & 28.3 & \textbf{42.4} & 62.0 & 37.2 & 22.7 & 33.9 \\
DINO-B & \mmedium & latent & 67.3 & 41.6 & 26.8 & 35.2 & 69.6 & 41.8 & 27.4 & 33.8 \\
DINO-B & \mmedium & program & 61.4 & 38.0 & 23.6 & 37.1 & 62.0 & 38.0 & 24.0 & 35.6 \\
\cmidrule(l){1-11}
DINO-L & \mmedium & latent & 65.6 & 41.4 & 27.4 & 34.2 & \textbf{71.3} & \textbf{41.9} & \textbf{29.0} & 35.8 \\
DINO-L & \mmedium & program & 60.3 & 37.0 & 24.6 & 37.6 & 64.0 & 38.5 & 25.9 & 37.6 \\
\bottomrule
\end{tabular*}
\end{table}

\begin{table}[t]
\centering
\caption{Monocular depth estimation on NYUv2 for all model variants, trained on COCO or ImageNet. All \mname rows and the matched teacher rows use the $256{\times}256$ protocol of Appendix~\ref{app:evaluation_protocol} (final checkpoint); RMSE in meters, AbsRel and $\delta_1$ in \%, computed over valid depth pixels. Teacher rows probe frozen DINO patch features; the bottom block evaluates the teachers with the official DINOv3 depth protocol as a reference and is not comparable with the matched rows. Latent uses the interpreter's spatial latent field; program uses the active program token vectors $z_q$. The best non-teacher value in each column is \textbf{bolded}.}
\label{tab:depth-coco}
\setlength{\tabcolsep}{3pt}
\renewcommand{\arraystretch}{1.1}
\footnotesize
\begin{tabular*}{\linewidth}{@{\extracolsep{\fill}}lll rrr rrr@{}}
\toprule
 & & & \multicolumn{3}{c}{Trained on COCO} & \multicolumn{3}{c}{Trained on ImageNet} \\
\cmidrule(lr){4-6}\cmidrule(lr){7-9}
Encoder & Model & Repr. & RMSE$\downarrow$ & AbsRel$\downarrow$ & $\delta_1\uparrow$ & RMSE$\downarrow$ & AbsRel$\downarrow$ & $\delta_1\uparrow$ \\
\midrule
\multicolumn{9}{l}{\emph{Teacher (frozen DINO), matched protocol}} \\
DINO-S & --- & teacher & 0.481 & 14.7 & 80.7 & 0.481 & 14.7 & 80.7 \\
DINO-B & --- & teacher & 0.424 & 12.8 & 85.1 & 0.424 & 12.8 & 85.1 \\
DINO-L & --- & teacher & 0.382 & 11.2 & 89.0 & 0.382 & 11.2 & 89.0 \\
\midrule
\multicolumn{9}{l}{\emph{\mname (ours)}} \\
DINO-S & \msmall & latent & 0.562 & 17.4 & 75.1 & 0.570 & 17.7 & 73.6 \\
DINO-S & \msmall & program & 0.559 & 17.2 & 76.0 & 0.572 & 16.8 & 75.1 \\
DINO-S & \mmedium & latent & 0.592 & 19.0 & 72.5 & 0.565 & 17.5 & 74.6 \\
DINO-S & \mmedium & program & 0.583 & 17.9 & 74.3 & 0.578 & 17.1 & 74.8 \\
\cmidrule(l){1-9}
DINO-B & \msmall & latent & 0.527 & 16.9 & 76.3 & 0.571 & 17.3 & 74.6 \\
DINO-B & \msmall & program & 0.525 & 15.3 & 79.1 & 0.567 & 16.8 & 75.7 \\
DINO-B & \mmedium & latent & 0.583 & 18.5 & 73.8 & \textbf{0.546} & \textbf{16.1} & \textbf{77.2} \\
DINO-B & \mmedium & program & 0.573 & 17.4 & 76.0 & 0.568 & 16.3 & 76.1 \\
\cmidrule(l){1-9}
DINO-L & \mmedium & latent & 0.532 & 16.0 & 77.9 & 0.554 & 17.0 & 76.4 \\
DINO-L & \mmedium & program & \textbf{0.522} & \textbf{15.2} & \textbf{79.5} & 0.560 & 16.3 & 77.0 \\
\midrule
\multicolumn{9}{l}{\emph{Teacher, official DINOv3 protocol (reference)}} \\
DINO-S & --- & teacher & 0.405 & 12.0 & 87.3 & & & \\
DINO-B & --- & teacher & 0.372 & 11.3 & 89.6 & & & \\
DINO-L & --- & teacher & 0.351 & 10.2 & 91.9 & & & \\
\bottomrule
\end{tabular*}
\end{table}

\begin{figure}[t]
    \centering
    \includegraphics[width=0.8\textwidth]{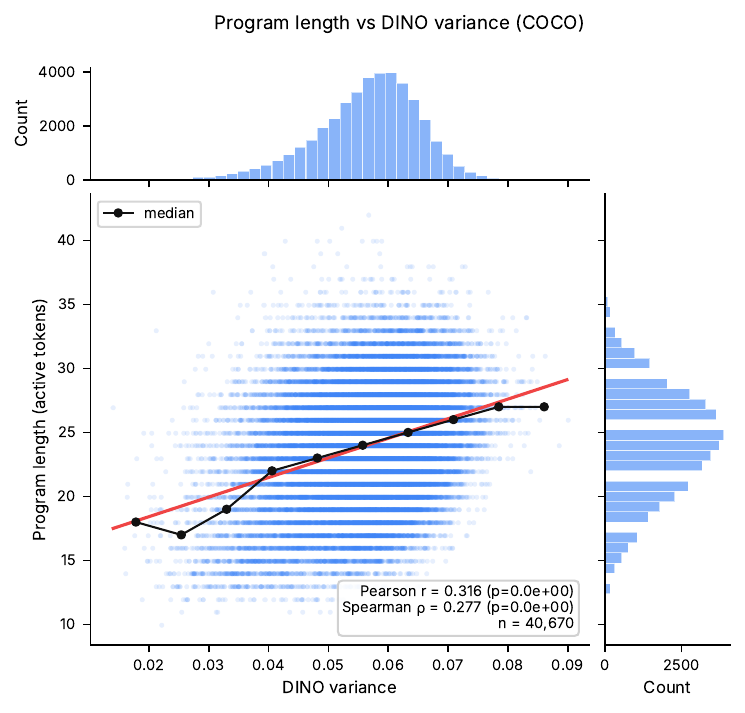}
    \caption{Program length and scene complexity analysis on COCO using variance of DINO embeddings as a proxy for scene complexity.}
    \label{fig:dino_complexity}
\end{figure}

\begin{figure}
    \centering
    \includegraphics[width=\linewidth]{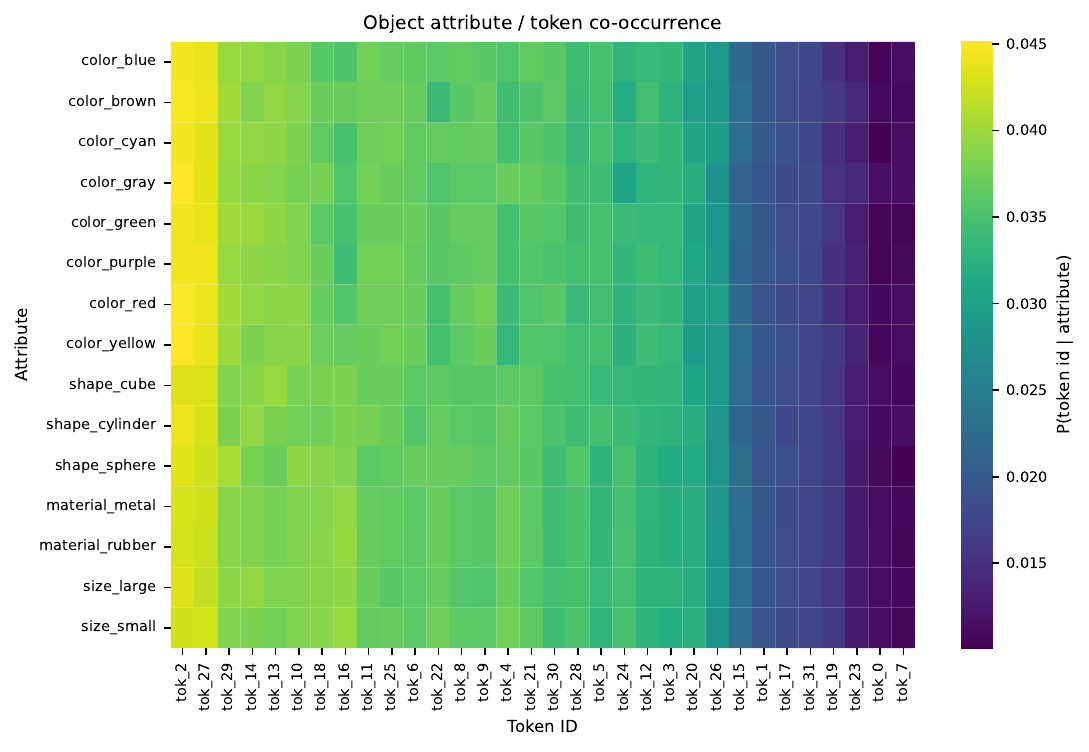}
    \caption{Contingency heatmap for CLEVR attributes and token ids; codebook size 32.}
    \label{fig:clevr_contingency}
\end{figure}

\FloatBarrier
\section{CLEVR}\label{app:clevr}

We train additional variants on CLEVR~\citep{clevr2017} at $128{\times}128$ resolution to study codebook sizing and program structure on a controlled scene domain. All CLEVR models use a DINO-S encoder with \mname-S generator/interpreter ($d_{\text{model}}{=}256$, 8 layers), and $K{=}32$ program tokens. The only variable is codebook size $V \in \{32, 64, 128\}$; all other hyperparameters match the main configuration (Table~\ref{tab:training}). Training runs for $5{\times}10^{5}$ steps with batch size 128 on a single H100 (${\sim}27$h each).

\begin{table}[t]
\centering
\small
\caption{CLEVR codebook-size variants (DINO-S / \mname-S, $K{=}32$, 128$\times$128).}
\label{tab:clevr_variants}
\begin{tabular}{c cccc c}
\toprule
$V$ & Cos$\uparrow$ & $R^2$$\uparrow$ & SSIM$\uparrow$ & LPIPS$\downarrow$ & V\% \\
\midrule
32  & .971 & .943 & .775 & .173 & 73.5 \\
64  & .965 & .931 & .756 & .187 & \textbf{94.7} \\
128 & \textbf{.973} & \textbf{.948} & \textbf{.787} & \textbf{.162} & 77.8 \\
\bottomrule
\end{tabular}
\end{table}

Table~\ref{tab:clevr_variants} reports results. All three codebook sizes achieve $R^2 > 0.93$, confirming that the architecture can faithfully represent CLEVR scenes through a short discrete program. The $V{=}128$ variant reaches the highest $R^2$ (.948) and SSIM (.787). Notably, $V{=}64$ shows a dip in alignment ($R^2{=}.931$) that is explained by its truncated codebook utilization equal to 95\%, meaning nearly every code is active within the adaptive prefix and the vocabulary has little remaining capacity. At $V{=}32$ the codebook is larger relative to scene complexity, giving more headroom (91\%), while at $V{=}128$ the vocabulary is comfortably under-utilized (86\%).

\textbf{Curriculum vs.\ fixed-length programs.}
To isolate the cost of adaptive-length training, we compare the curriculum variants ($V \in \{64, 128\}$) against matched fixed-length baselines that always use all $K{=}32$ tokens with no halting mechanism.

\begin{table}
\centering
\small
\caption{Curriculum vs.\ fixed-length training on CLEVR (DINOv3-S / \mname-S, $K{=}32$, 128$\times$128, 500k steps). Fixed-length models use all 32 tokens; curriculum models learn an adaptive prefix via the four-phase schedule (Sec.~\ref{sec:curriculum}).}
\label{tab:clevr_curr_vs_fixed}
\begin{tabular}{ll cc c}
\toprule
$V$ & Mode & $R^2$$\uparrow$ & V\% & Time (h) \\
\midrule
64  & Fixed      & \textbf{.961} & 88.6 & \textbf{24.7} \\
64  & Curriculum & .931 & 94.7 & 27.3 \\
\midrule
128 & Fixed      & \textbf{.966} & 94.2 & \textbf{24.9} \\
128 & Curriculum & .948 & 77.8 & 27.3 \\
\bottomrule
\end{tabular}
\end{table}

Table~\ref{tab:clevr_curr_vs_fixed} shows that fixed-length programs outperform curriculum variants on $R^2$, with gains of $.018$--$.030$. Fixed-length training is also ${\sim}10\%$ faster in wall time (24.7--24.9h vs.\ 27.3h) because it avoids the halting head forward pass, oracle target computation, and phase-dependent loss modulation. This gap is expected: the curriculum solves a strictly harder problem---learning \emph{when to stop} in addition to \emph{what to encode}---while the fixed-length model always has access to all $K$ tokens.

The curriculum's value is not in raw metric improvement but in the capability it unlocks: variable-length programs that allocate fewer tokens to simple scenes and more to complex ones (see Fig.~\ref{fig:clevr_main}). On CLEVR, where scene complexity varies from 3 to 10 objects, learned program lengths range from 8 to 32 tokens, enabling a ${\sim}2.5{\times}$ compression ratio on the simplest scenes with no manual tuning.

\begin{figure}[t]
    \centering
    \includegraphics[width=\linewidth]{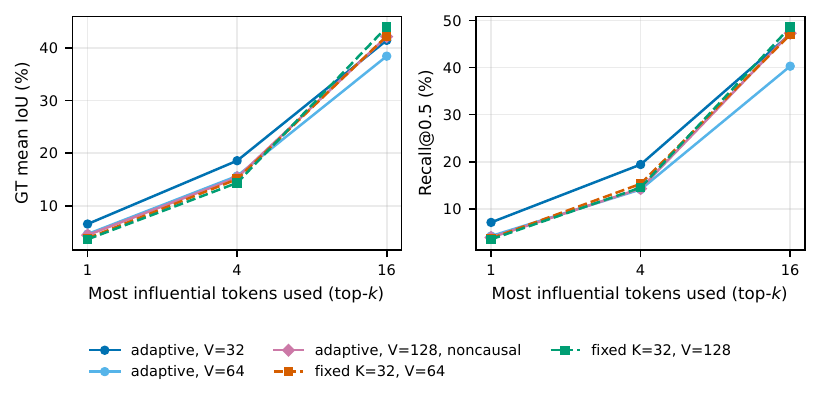}
    \caption{\textbf{Token grounding on CLEVR.} Overlap between the influence maps of the $k$ most influential program tokens and the ground-truth object masks on 5,000 CLEVR test images, for adaptive and fixed-length programs (protocol in Appendix~\ref{app:evaluation_protocol}). Left: mean IoU of each object with its matched token; right: fraction of objects matched with IoU $\geq 0.5$. Comparisons are relative: no random-token baseline is reported.}
    \label{fig:clevr-grounding}
\end{figure}

\textbf{Symbolic content and grounding.}
Table~\ref{tab:clevr-readouts} tests which scene properties can be read linearly from the discrete tokens of the CLEVR models, and Figure~\ref{fig:clevr-grounding} measures how well individual tokens localize objects. Linear readouts of the CLEVR tokens predict object count and attributes better when they are given token positions, and shuffling the token order removes that advantage (Table~\ref{tab:clevr-readouts}). On these symbolic targets, fixed-length programs are as informative as adaptive ones. The most influential tokens overlap individual objects, and the overlap grows as more tokens are included (Fig.~\ref{fig:clevr-grounding}).

\begin{table}[h]
\centering
\caption{What the discrete tokens encode on CLEVR. Linear predictors of object count (MAE, $\downarrow$) and of 39 attribute-presence labels (mean accuracy in \%, $\uparrow$) are trained on the tokens of frozen CLEVR models (DINO-S / \msmall, $128{\times}128$, programs of up to 32 codes; 90k / 5k / 5k train / validation / test images; epoch selected on validation, test reported). Bag uses code counts only; Bigram adds counts of adjacent code pairs; Pos.\ adds code-by-position indicators; Ord.\ combines both; Shuf.\ is Ord.\ with the active tokens randomly permuted. The best readout in each row and metric group is \textbf{bolded}: position-aware readouts are best for every model, and shuffling the order removes their advantage. Fixed-length programs are as informative as adaptive ones on these symbolic targets.}
\label{tab:clevr-readouts}
\setlength{\tabcolsep}{3pt}
\renewcommand{\arraystretch}{1.1}
\footnotesize
\begin{tabular*}{\linewidth}{@{\extracolsep{\fill}}l rrrrr rrrrr@{}}
\toprule
 & \multicolumn{5}{c}{Object count MAE$\downarrow$} & \multicolumn{5}{c}{Attribute accuracy (\%)$\uparrow$} \\
\cmidrule(lr){2-6}\cmidrule(lr){7-11}
Model & Bag & Bigram & Pos. & Ord. & Shuf. & Bag & Bigram & Pos. & Ord. & Shuf. \\
\midrule
Adaptive, $V{=}32$ & 0.943 & 0.900 & 0.705 & \textbf{0.700} & 0.956 & 76.4 & 76.1 & \textbf{78.0} & 77.6 & 75.5 \\
Adaptive, $V{=}64$ & 1.063 & 1.025 & \textbf{0.742} & 0.764 & 1.083 & 76.2 & 74.4 & \textbf{76.5} & 75.2 & 74.5 \\
Adaptive, $V{=}128$, noncausal & 0.932 & 0.928 & \textbf{0.686} & 0.721 & 0.974 & 76.4 & 75.2 & \textbf{78.6} & 77.5 & 74.3 \\
\midrule
Fixed $K{=}32$, $V{=}64$ & 0.876 & 0.851 & \textbf{0.613} & 0.629 & 0.926 & 76.6 & 76.1 & \textbf{80.9} & 80.1 & 74.7 \\
Fixed $K{=}32$, $V{=}128$ & 0.907 & 0.916 & \textbf{0.574} & 0.634 & 1.004 & 76.9 & 75.7 & \textbf{81.7} & 80.3 & 74.0 \\
\bottomrule
\end{tabular*}
\end{table}

\FloatBarrier
\section{LLM Scene Decoding from Learned Visual Tokens}
\label{app:llm_scene_decoding}

We evaluate whether frontier large language models can decode continuous CLEVR scene descriptions directly from discrete codebook tokens produced by a learned visual tokenizer. Given a sequence of integer codes from a DINO-S encoder (with codebook sizes $V \in \{32, 64, 128\}$), the models must predict the full scene structure as a JSON string, including object count, their attributes and continuous $(x, y)$ positions. We benchmark three frontier LLMs (Qwen3.5-397B-A17B, DeepSeek-V3.1-vLLM-2, and GPT-OSS-120B, all from Hugging Face) against a \textit{Naive Baseline}, which uses no language model and predicts a fixed ``average scene'' derived entirely from the statistical distributions of the few-shot pool. This baseline establishes a lower bound measuring how much of a scene can be guessed from prior dataset statistics alone, without decoding the actual visual tokens. All models are evaluated on 15 held-out test images using a 50-shot random in-context learning setup, with one retry permitted on parse failure.

\begin{table}[t]
\centering
\caption{Scene decoding performance of frontier LLMs compared to a statistical baseline. Parse Fails denote outputs with invalid JSON structures (max one retry). Metrics are averaged across 15 test examples. Best LLM result per metric in \textbf{bold}.}
\label{tab:llm_scene_decoding}
\small
\begin{tabular}{lcccccc}
\toprule
Model & $V$ & Parse Fails & Count Err $\downarrow$ & Attr Acc $\uparrow$ & Pos MSE $\downarrow$ & Pos MAE $\downarrow$ \\
\midrule
Naive Baseline & N/A & 0 / 15 & 2.27 & 0.568 & 3.84 & 2.22 \\
\midrule
GPT-OSS-120B & 32 & 1 / 15 & 2.36 & 0.469 & 2.45 & 1.71 \\
GPT-OSS-120B & 64 & 3 / 15 & 2.83 & 0.536 & 3.26 & 1.87 \\
GPT-OSS-120B & 128 & 1 / 15 & 2.36 & 0.524 & 4.61 & 2.13 \\
\midrule
Qwen3.5-397B-A17B & 32 & 0 / 15 & 1.43 & 0.510 & 3.22 & 1.90 \\
Qwen3.5-397B-A17B & 64 & 0 / 15 & 1.00 & 0.458 & 3.37 & 2.00 \\
Qwen3.5-397B-A17B & 128 & 0 / 15 & 1.53 & 0.505 & 3.17 & 1.91 \\
\midrule
DeepSeek-V3.1-vLLM-2 & 32 & 1 / 15 & 1.36 & 0.513 & 3.68 & 2.11 \\
DeepSeek-V3.1-vLLM-2 & 64 & 0 / 15 & \textbf{0.93} & \textbf{0.543} & 3.64 & 1.97 \\
DeepSeek-V3.1-vLLM-2 & 128 & 0 / 15 & 1.13 & 0.487 & \textbf{3.15} & \textbf{1.84} \\
\bottomrule
\end{tabular}
\end{table}

\paragraph{Key findings.} Two observations stand out. First, the stronger frontier models (Qwen3.5 and DeepSeek) decode \emph{spatial} structure substantially better than the naive baseline across all codebook sizes, with DeepSeek achieving the best count error ($0.93$ at $V{=}64$) and the best position errors ($3.15$ MSE / $1.84$ MAE at $V{=}128$). Second, and more strikingly, \emph{no model surpasses the naive baseline's attribute accuracy of $56.8\%$}: while the models recover ``where'' and ``how many'' (Count Err), binding tokens to discrete textual properties such as color or material remains beyond their reach without dataset priors. The smaller GPT-OSS-120B exhibits a different failure mode entirely, with count errors ($\sim 2.4$--$2.8$) at or above the naive baseline despite achieving competitive position errors at $V{=}32$, suggesting it leans more heavily on memorized layout statistics than on token-level decoding.

\paragraph{Attribute--token co-occurrence.}
Figure~\ref{fig:clevr_contingency} shows the contingency table between CLEVR object attributes (color, shape, material, size) and assigned token IDs. The heatmap reveals no strong systematic correspondence---consistent with the observation that attribute identity is not yet reliably encoded in individual codebook entries. Understanding whether attribute information is distributed across token combinations rather than localized in single tokens is an important direction for future work.

We document the exact prompt used to elicit scene predictions from each frontier LLM. All models were queried at temperature $0.0$ with 50 random in-context examples preceding the test query. To minimize token consumption across the 50-shot context, attribute values are emitted in an abbreviated form, with a per-prompt legend instructing the model to use the short codes.

\paragraph{Attribute abbreviations.} Object attributes are remapped to single- or two-character codes, summarized in Table~\ref{tab:attr_abbr}. The same scheme is applied uniformly to the few-shot examples, the schema description in the prompt, and the expected model output, so that the model never needs to translate between long and short forms.

\begin{table}[t]
\centering
\caption{Attribute abbreviation scheme used in both few-shot examples and model outputs.}
\label{tab:attr_abbr}
\small
\begin{tabular}{llll}
\toprule
\textbf{Color} & \textbf{Shape} & \textbf{Size} & \textbf{Material} \\
\midrule
gray $\rightarrow$ \texttt{gr}     & cube $\rightarrow$ \texttt{cu}     & small $\rightarrow$ \texttt{S} & rubber $\rightarrow$ \texttt{R} \\
red $\rightarrow$ \texttt{r}       & sphere $\rightarrow$ \texttt{sp}   & large $\rightarrow$ \texttt{L} & metal  $\rightarrow$ \texttt{M} \\
blue $\rightarrow$ \texttt{bl}     & cylinder $\rightarrow$ \texttt{cy} &                                &                                 \\
green $\rightarrow$ \texttt{gn}    &                                    &                                &                                 \\
brown $\rightarrow$ \texttt{br}    &                                    &                                &                                 \\
purple $\rightarrow$ \texttt{pr}   &                                    &                                &                                 \\
cyan $\rightarrow$ \texttt{cy}     &                                    &                                &                                 \\
yellow $\rightarrow$ \texttt{y}    &                                    &                                &                                 \\
\bottomrule
\end{tabular}
\end{table}

\paragraph{Prompt template.} The prompt below is templated with the codebook size $V$ (substituted for \texttt{\{codebook\_size\}}), the maximum code index $V-1$ (substituted for \texttt{\{codebook\_max\}}), the inline abbreviation legend (\texttt{\{abbr\_legend\}}), and an optional reasoning suffix (\texttt{\{think\_suffix\}}) used only for DeepSeek-V3.1 to elicit step-by-step reasoning prior to JSON emission.

\begin{tcolorbox}[
  colback=gray!5,
  colframe=gray!50,
  title=\textbf{Decoding prompt},
  fonttitle=\bfseries,
  boxrule=0.5pt,
  arc=2pt
]
\small\ttfamily
Predict a CLEVR scene (3--10 objects) from integer codes produced by a visual tokenizer (codebook size=\{codebook\_size\}, codes in [0,\{codebook\_max\}]).\\[2pt]
Output compact JSON using short keys and abbreviated values:\\
\{abbr\_legend\}\\[2pt]
Schema: \{"o":[\{"s":"<shape>","c":"<color>","z":"<size>",\\"m":"<mat>","p":[x,y]\},...]\}\\[2pt]
Position: x,y in [-3,3], 2 decimal places.\\[2pt]
Example: \{"o":[\{"s":"cu","c":"r","z":"L","m":"M","p":[1.23,-0.45]\}]\}\\[2pt]
JSON only. No prose.\{think\_suffix\}
\end{tcolorbox}

\paragraph{Reasoning suffix.} For DeepSeek-V3.1 and GPT-OSS-120B, the placeholder \texttt{\{think\_suffix\}} is replaced with the following string; for Qwen3.5 it is left empty.

\begin{tcolorbox}[
  colback=gray!5,
  colframe=gray!50,
  title=\textbf{DeepSeek reasoning suffix},
  fonttitle=\bfseries,
  boxrule=0.5pt,
  arc=2pt
]
\small\ttfamily
Before your final answer, reason step-by-step inside <think> and </think> tags. Then output the JSON.
\end{tcolorbox}

\paragraph{Decoding settings.} All models were queried with temperature $0.0$ to encourage deterministic outputs. A single retry was permitted on JSON parse failure, after which the example was counted as a parse failure in Table~\ref{tab:llm_scene_decoding}.

%% file: body/figure-segmentation-probes-coco.tex
\begin{figure}
    \centering
    \includegraphics[width=1\linewidth]{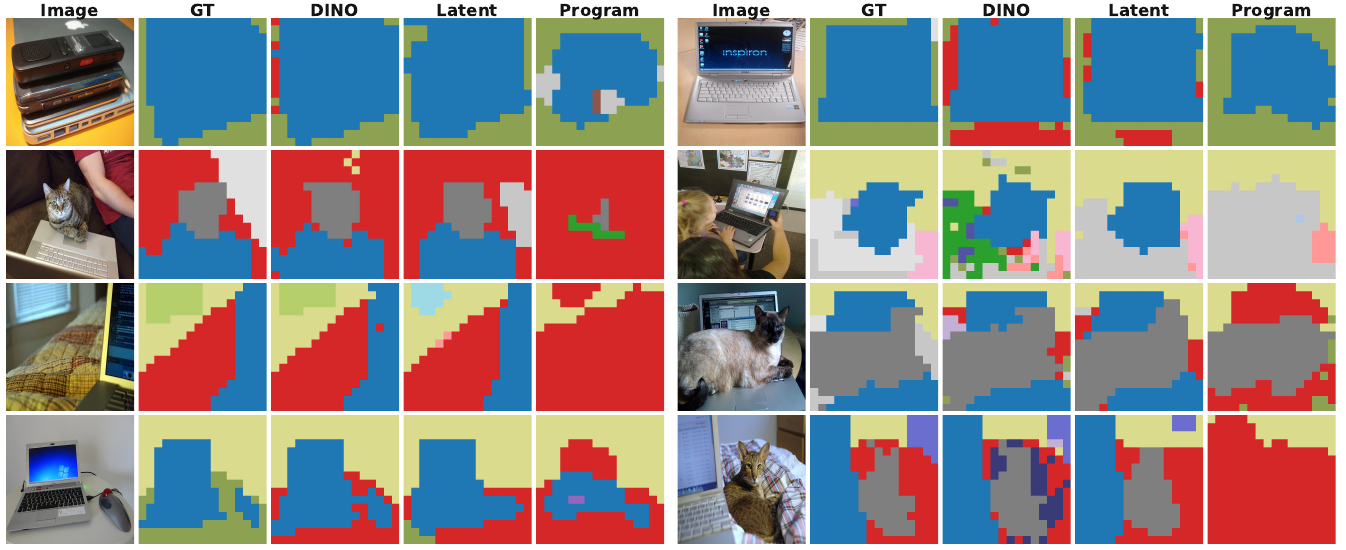}
    \caption{Supervised downstream segmentation probes on COCO-Stuff-27. DINO and Latent show predictions from linear patch probes trained on frozen DINO teacher features and \mname latent patch features, respectively. Program shows predictions from a probe trained on the
  quantized program.
}
    \label{fig:downstream_segmentation_prob_grid}
\end{figure}

%% file: references.bib
@inproceedings{vandenoord2017vqvae,
  title     = {Neural Discrete Representation Learning},
  author    = {van den Oord, A{\"a}ron and Vinyals, Oriol and Kavukcuoglu, Koray},
  booktitle = {Advances in Neural Information Processing Systems (NeurIPS)},
  year      = {2017}
}

@inproceedings{razavi2019vqvae2,
  title     = {Generating Diverse High-Fidelity Images with {VQ-VAE-2}},
  author    = {Razavi, Ali and van den Oord, A{\"a}ron and Vinyals, Oriol},
  booktitle = {Advances in Neural Information Processing Systems (NeurIPS)},
  year      = {2019}
}

@inproceedings{esser2021vqgan,
  title     = {Taming Transformers for High-Resolution Image Synthesis},
  author    = {Esser, Patrick and Rombach, Robin and Ommer, Bj{\"o}rn},
  booktitle = {IEEE/CVF Conference on Computer Vision and Pattern Recognition (CVPR)},
  year      = {2021}
}

@inproceedings{chang2022maskgit,
  title     = {{MaskGIT}: Masked Generative Image Transformer},
  author    = {Chang, Huiwen and Zhang, Han and Jiang, Lu and Liu, Ce and Freeman, William T.},
  booktitle = {IEEE/CVF Conference on Computer Vision and Pattern Recognition (CVPR)},
  year      = {2022}
}

@inproceedings{tian2024var,
  title     = {Visual Autoregressive Modeling: Scalable Image Generation via Next-Scale Prediction},
  author    = {Tian, Keyu and Jiang, Yi and Yuan, Zehuan and Peng, Bingyue and Wang, Liwei},
  booktitle = {Advances in Neural Information Processing Systems (NeurIPS)},
  year      = {2024},
  eprint    = {2404.02905},
  archivePrefix = {arXiv}
}

@inproceedings{mentzer2024fsq,
  title     = {Finite Scalar Quantization: {VQ-VAE} Made Simple},
  author    = {Mentzer, Fabian and Minnen, David and Agustsson, Eirikur and Tschannen, Michael},
  booktitle = {International Conference on Learning Representations (ICLR)},
  year      = {2024},
  eprint    = {2309.15505},
  archivePrefix = {arXiv}
}

@inproceedings{lee2022rqvae,
  title     = {Autoregressive Image Generation Using Residual Quantization},
  author    = {Lee, Doyup and Kim, Chiheon and Kim, Saehoon and Cho, Minsu and Han, Wook-Shin},
  booktitle = {IEEE/CVF Conference on Computer Vision and Pattern Recognition (CVPR)},
  year      = {2022}
}

@inproceedings{fifty2024rotation,
  title     = {Restructuring Vector Quantization with the Rotation Trick},
  author    = {Fifty, Christopher and Junkins, Ronald G. and Duan, Dennis and Iger, Aniketh and Liu, Jerry W. and Amid, Ehsan and Thrun, Sebastian and R{\'e}, Christopher},
  booktitle = {International Conference on Learning Representations (ICLR)},
  year      = {2025},
  eprint    = {2410.06424},
  archivePrefix = {arXiv}
}

@inproceedings{chen2024softvq,
  title     = {{SoftVQ-VAE}: Efficient 1-Dimensional Continuous Tokenizer},
  author    = {Chen, Hao and Wang, Ze and Li, Xiang and Sun, Ximeng and Chen, Fangyi and Liu, Jiang and Wang, Jindong and Raj, Bhiksha and Liu, Zicheng and Barsoum, Emad},
  booktitle = {IEEE/CVF Conference on Computer Vision and Pattern Recognition (CVPR)},
  year      = {2025},
  eprint    = {2412.10958},
  archivePrefix = {arXiv}
}

@inproceedings{li2024imagefolder,
  title     = {{ImageFolder}: Autoregressive Image Generation with Folded Tokens},
  author    = {Li, Xiang and Qiu, Kai and Chen, Hao and Kuen, Jason and Gu, Jiuxiang and Raj, Bhiksha and Lin, Zhe},
  year      = {2024},
  eprint    = {2410.01756},
  archivePrefix = {arXiv}
}

@inproceedings{yu2024titok,
  title     = {An Image is Worth 32 Tokens for Reconstruction and Generation},
  author    = {Yu, Qihang and Weber, Mark and Deng, Xueqing and Shen, Xiaohui and Cremers, Daniel and Chen, Liang-Chieh},
  booktitle = {Advances in Neural Information Processing Systems (NeurIPS)},
  year      = {2024},
  eprint    = {2406.07550},
  archivePrefix = {arXiv}
}

@inproceedings{bachmann2025flextok,
  title     = {{FlexTok}: Resampling Images into 1{D} Token Sequences of Flexible Length},
  author    = {Bachmann, Roman and Allardice, Jesse and Mizrahi, David and Fini, Enrico and Kar, O{\u{g}}uzhan Fatih and Amirloo, Elmira and El-Nouby, Alaaeldin and Zamir, Amir and Dehghan, Afshin},
  booktitle = {International Conference on Machine Learning (ICML)},
  year      = {2025},
  eprint    = {2502.13967},
  archivePrefix = {arXiv}
}

@article{miwa2025onedpiece,
  title   = {{One-D-Piece}: Image Tokenizer Meets Quality-Controllable Compression},
  author  = {Miwa, Keita and Sasaki, Kento and Arai, Hidehisa and Takahashi, Tsubasa and Yamaguchi, Yu},
  journal = {arXiv preprint arXiv:2501.10064},
  year    = {2025}
}

@inproceedings{duggal2025alit,
  title     = {Adaptive Length Image Tokenization via Recurrent Allocation},
  author    = {Duggal, Shivam and Isola, Phillip and Torralba, Antonio and Freeman, William T.},
  booktitle = {International Conference on Learning Representations (ICLR)},
  year      = {2025},
  eprint    = {2411.02393},
  archivePrefix = {arXiv}
}

@inproceedings{yan2025elastictok,
  title     = {{ElasticTok}: Adaptive Tokenization for Image and Video},
  author    = {Yan, Wilson and Zaharia, Matei and Mnih, Volodymyr and Abbeel, Pieter and Faust, Aleksandra and Liu, Hao},
  booktitle = {International Conference on Learning Representations (ICLR)},
  year      = {2025},
  eprint    = {2410.08368},
  archivePrefix = {arXiv}
}

@article{shen2025cat,
  title   = {{CAT}: Content-Adaptive Image Tokenization},
  author  = {Shen, Junhong and Tirumala, Kushal and Yasunaga, Michihiro and Misra, Ishan and Zettlemoyer, Luke and Yu, Lili and Zhou, Chunting},
  journal = {arXiv preprint arXiv:2501.03120},
  year    = {2025}
}

@article{ye2025infotok,
  title   = {{InfoTok}: Adaptive Discrete Video Tokenizer via Information-Theoretic Compression},
  author  = {Ye, Haotian and He, Qiyuan and Han, Jiaqi and Li, Puheng and Fan, Jiaojiao and Hao, Zekun and Reda, Fitsum and Balaji, Yogesh and Chen, Huayu and Liu, Sheng and Yao, Angela and Zou, James and Ermon, Stefano and Wang, Haoxiang and Liu, Ming-Yu},
  journal = {arXiv preprint arXiv:2512.16975},
  year    = {2025}
}

@article{peng2022beitv2,
  title   = {{BEiT v2}: Masked Image Modeling with Vector-Quantized Visual Tokenizers},
  author  = {Peng, Zhiliang and Dong, Li and Bao, Hangbo and Ye, Qixiang and Wei, Furu},
  journal = {arXiv preprint arXiv:2208.06366},
  year    = {2022}
}

@inproceedings{li2024rcg,
  title     = {Return of Unconditional Generation: A Self-Supervised Representation Generation Method},
  author    = {Li, Tianhong and Katabi, Dina and He, Kaiming},
  booktitle = {Advances in Neural Information Processing Systems (NeurIPS)},
  year      = {2024},
  eprint    = {2312.03701},
  archivePrefix = {arXiv}
}

@inproceedings{yu2025repa,
  title     = {Representation Alignment for Generation: Training Diffusion Transformers Is Easier Than You Think},
  author    = {Yu, Sihyun and Kwak, Sangkyung and Jang, Huiwon and Jeong, Jongheon and Huang, Jonathan and Shin, Jinwoo and Xie, Saining},
  booktitle = {International Conference on Learning Representations (ICLR)},
  year      = {2025},
  eprint    = {2410.06940},
  archivePrefix = {arXiv}
}

@inproceedings{leng2025repae,
  title     = {{REPA-E}: Unlocking {VAE} for End-to-End Tuning with Latent Diffusion Transformers},
  author    = {Leng, Xingjian and Singh, Jaskirat and Hou, Yunzhong and Xing, Zhenchang and Xie, Saining and Zheng, Liang},
  booktitle = {IEEE/CVF International Conference on Computer Vision (ICCV)},
  year      = {2025},
  eprint    = {2504.10483},
  archivePrefix = {arXiv}
}

@inproceedings{yao2025vavae,
  title     = {Reconstruction vs.\ Generation: Taming Optimization Dilemma in Latent Diffusion Models},
  author    = {Yao, Jingfeng and Wang, Xinggang},
  booktitle = {IEEE/CVF Conference on Computer Vision and Pattern Recognition (CVPR)},
  year      = {2025},
  eprint    = {2501.01423},
  archivePrefix = {arXiv}
}

@inproceedings{chen2025maetok,
  title     = {Masked Autoencoders Are Effective Tokenizers for Diffusion Models},
  author    = {Chen, Hao and Han, Yujin and Chen, Fangyi and Li, Xiang and Wang, Yidong and Wang, Jindong and Wang, Ze and Liu, Zicheng and Zou, Difan and Raj, Bhiksha},
  booktitle = {International Conference on Machine Learning (ICML)},
  year      = {2025},
  eprint    = {2502.03444},
  archivePrefix = {arXiv}
}

@article{yang2025ldetok,
  title   = {Latent Denoising Makes Good Visual Tokenizers},
  author  = {Yang, Jiawei and Li, Tianhong and Fan, Lijie and Tian, Yonglong and Wang, Yue},
  journal = {arXiv preprint arXiv:2507.15856},
  year    = {2025}
}

@inproceedings{wen2025principal,
  title     = {``{P}rincipal Components'' Enable a New Language of Images},
  author    = {Wen, Xin and Zhao, Bingchen and Elezi, Ismail and Deng, Jiankang and Qi, Xiaojuan},
  booktitle = {IEEE/CVF International Conference on Computer Vision (ICCV)},
  year      = {2025},
  eprint    = {2503.08685},
  archivePrefix = {arXiv}
}

@article{simeoni2025dinov3,
  title   = {{DINOv3}},
  author  = {Sim{\'e}oni, Oriane and Vo, Huy V. and Seitzer, Maximilian and Baldassarre, Federico and Oquab, Maxime and Jose, Cijo and Khalidov, Vasil and Szafraniec, Marc and Yi, Seungeun and Ramamonjisoa, Micha{\"e}l and Massa, Francisco and Haziza, Daniel and Wehrstedt, Luca and Wang, Jianyuan and Darcet, Timoth{\'e}e and Moutakanni, Th{\'e}o and Sentana, Leonel and Roberts, Claire and Vedaldi, Andrea and Tolan, Jamie and Brandt, John and Couprie, Camille and Mairal, Julien and J{\'e}gou, Herv{\'e} and Labatut, Patrick and Bojanowski, Piotr},
  journal = {arXiv preprint arXiv:2508.10104},
  year    = {2025}
}

@inproceedings{locatello2020slot,
  title     = {Object-Centric Learning with Slot Attention},
  author    = {Locatello, Francesco and Weissenborn, Dirk and Unterthiner, Thomas and Mahendran, Aravindh and Heigold, Georg and Uszkoreit, Jakob and Dosovitskiy, Alexey and Kipf, Thomas},
  booktitle = {Advances in Neural Information Processing Systems (NeurIPS)},
  year      = {2020}
}

@inproceedings{singh2022slate,
  title     = {Illiterate {DALL-E} Learns to Compose},
  author    = {Singh, Gautam and Deng, Fei and Ahn, Sungjin},
  booktitle = {International Conference on Learning Representations (ICLR)},
  year      = {2022},
  eprint    = {2110.11405},
  archivePrefix = {arXiv}
}

@inproceedings{seitzer2023dinosaur,
  title     = {Bridging the Gap to Real-World Object-Centric Learning},
  author    = {Seitzer, Maximilian and Horn, Max and Zadaianchuk, Andrii and Zietlow, Dominik and Xiao, Tianjun and Simon-Gabriel, Carl-Johann and He, Tong and Zhang, Zheng and Sch{\"o}lkopf, Bernhard and Brox, Thomas and Locatello, Francesco},
  booktitle = {International Conference on Learning Representations (ICLR)},
  year      = {2023},
  eprint    = {2209.14860},
  archivePrefix = {arXiv}
}

@inproceedings{wu2023slotdiffusion,
  title     = {{SlotDiffusion}: Object-Centric Generative Modeling with Diffusion Models},
  author    = {Wu, Ziyi and Hu, Jingyu and Lu, Wuyue and Gilitschenski, Igor and Garg, Animesh},
  booktitle = {Advances in Neural Information Processing Systems (NeurIPS)},
  year      = {2023},
  eprint    = {2305.11281},
  archivePrefix = {arXiv}
}

@inproceedings{reed2016npi,
  title     = {Neural Programmer-Interpreters},
  author    = {Reed, Scott and de Freitas, Nando},
  booktitle = {International Conference on Learning Representations (ICLR)},
  year      = {2016},
  eprint    = {1511.06279},
  archivePrefix = {arXiv}
}

@inproceedings{carion2020detr,
  title     = {End-to-End Object Detection with Transformers},
  author    = {Carion, Nicolas and Massa, Francisco and Synnaeve, Gabriel and Usunier, Nicolas and Kirillov, Alexander and Zagoruyko, Sergey},
  booktitle = {European Conference on Computer Vision (ECCV)},
  year      = {2020}
}

@inproceedings{jaegle2021perceiver,
  title     = {Perceiver: General Perception with Iterative Attention},
  author    = {Jaegle, Andrew and Gimeno, Felix and Brock, Andrew and Zisserman, Andrew and Vinyals, Oriol and Carreira, Jo{\~a}o},
  booktitle = {International Conference on Machine Learning (ICML)},
  year      = {2021}
}

@inproceedings{jaegle2022perceiverio,
  title     = {Perceiver {IO}: A General Architecture for Structured Inputs \& Outputs},
  author    = {Jaegle, Andrew and Borgeaud, Sebastian and Alayrac, Jean-Baptiste and Doersch, Carl and Ionescu, Catalin and Ding, David and Koppula, Skanda and Zoran, Daniel and Brock, Andrew and Shelhamer, Evan and H{\'e}naff, Olivier and Botvinick, Matthew M. and Zisserman, Andrew and Vinyals, Oriol and Carreira, Jo{\~a}o},
  booktitle = {International Conference on Learning Representations (ICLR)},
  year      = {2022}
}

@INPROCEEDINGS{imagenet2009,
  author={Deng, Jia and Dong, Wei and Socher, Richard and Li, Li-Jia and Kai Li and Li Fei-Fei},
  booktitle={2009 IEEE Conference on Computer Vision and Pattern Recognition}, 
  title={ImageNet: A large-scale hierarchical image database}, 
  year={2009},
  volume={},
  number={},
  pages={248-255},
  doi={10.1109/CVPR.2009.5206848}}

@article{clevr2017,
  author       = {Justin Johnson and
                  Bharath Hariharan and
                  Laurens van der Maaten and
                  Li Fei{-}Fei and
                  C. Lawrence Zitnick and
                  Ross B. Girshick},
  title        = {{CLEVR:} {A} Diagnostic Dataset for Compositional Language and Elementary
                  Visual Reasoning},
  journal      = {CoRR},
  volume       = {abs/1612.06890},
  year         = {2016},
  url          = {http://arxiv.org/abs/1612.06890},
  eprinttype   = {arXiv},
  eprint       = {1612.06890},
  bibsource    = {dblp computer science bibliography, https://dblp.org}
}

@article{coco2014,
  author       = {Tsung{-}Yi Lin and
                  Michael Maire and
                  Serge J. Belongie and
                  Lubomir D. Bourdev and
                  Ross B. Girshick and
                  James Hays and
                  Pietro Perona and
                  Deva Ramanan and
                  Piotr Doll{\'{a}}r and
                  C. Lawrence Zitnick},
  title        = {Microsoft {COCO:} Common Objects in Context},
  journal      = {CoRR},
  volume       = {abs/1405.0312},
  year         = {2014},
  url          = {http://arxiv.org/abs/1405.0312},
  eprinttype   = {arXiv},
  eprint       = {1405.0312},
  bibsource    = {dblp computer science bibliography, https://dblp.org}
}

@misc{dvt-survey2026,
      title={From Principles to Applications: A Comprehensive Survey of Discrete Tokenizers in Generation, Comprehension, Recommendation, and Information Retrieval}, 
      author={Jian Jia and Jingtong Gao and Ben Xue and Junhao Wang and Qingpeng Cai and Quan Chen and Xiangyu Zhao and Peng Jiang and Kun Gai},
      year={2025},
      eprint={2502.12448},
      archivePrefix={arXiv},
      primaryClass={cs.IR},
      url={https://arxiv.org/abs/2502.12448}, 
}

@inproceedings{
yu2024magvitv2,
title={Language Model Beats Diffusion - Tokenizer is key to visual generation},
author={Lijun Yu and Jose Lezama and Nitesh Bharadwaj Gundavarapu and Luca Versari and Kihyuk Sohn and David Minnen and Yong Cheng and Agrim Gupta and Xiuye Gu and Alexander G Hauptmann and Boqing Gong and Ming-Hsuan Yang and Irfan Essa and David A Ross and Lu Jiang},
booktitle={The Twelfth International Conference on Learning Representations},
year={2024},
url={https://openreview.net/forum?id=gzqrANCF4g}
}

@misc{zhao2025bsq,
      title={Spherical Leech Quantization for Visual Tokenization and Generation}, 
      author={Yue Zhao and Hanwen Jiang and Zhenlin Xu and Chutong Yang and Ehsan Adeli and Philipp Krähenbühl},
      year={2025},
      eprint={2512.14697},
      archivePrefix={arXiv},
      primaryClass={cs.CV},
      url={https://arxiv.org/abs/2512.14697}, 
}

@inproceedings{
duggal2025karl,
title={Single-pass Adaptive Image Tokenization for Minimum Program Search},
author={Shivam Duggal and Sanghyun Byun and William T. Freeman and Antonio Torralba and Phillip Isola},
booktitle={The Thirty-ninth Annual Conference on Neural Information Processing Systems},
year={2025},
url={https://openreview.net/forum?id=HkTOnCUQ1z}
}

@INPROCEEDINGS {assran2023ijepa,
author = { Assran, Mahmoud and Duval, Quentin and Misra, Ishan and Bojanowski, Piotr and Vincent, Pascal and Rabbat, Michael and LeCun, Yann and Ballas, Nicolas },
booktitle = { 2023 IEEE/CVF Conference on Computer Vision and Pattern Recognition (CVPR) },
title = {{ Self-Supervised Learning from Images with a Joint-Embedding Predictive Architecture }},
year = {2023},
volume = {},
ISSN = {},
pages = {15619-15629},
doi = {10.1109/CVPR52729.2023.01499},
url = {https://doi.ieeecomputersociety.org/10.1109/CVPR52729.2023.01499},
publisher = {IEEE Computer Society},
address = {Los Alamitos, CA, USA},
month =Jun}

@inproceedings{
baek2026neo,
title={Learning to Theorize the World from Observation},
author={Doojin Baek and Gyubin Lee and Junyeob Baek and Hosung Lee and Sungjin Ahn},
booktitle={Forty-third International Conference on Machine Learning},
year={2026},
url={https://openreview.net/forum?id=wsA8LgHU5U}
}

@inproceedings{
wu2023slotformer,
title={SlotFormer: Unsupervised Visual Dynamics Simulation with Object-Centric Models},
author={Ziyi Wu and Nikita Dvornik and Klaus Greff and Thomas Kipf and Animesh Garg},
booktitle={The Eleventh International Conference on Learning Representations },
year={2023},
url={https://openreview.net/forum?id=TFbwV6I0VLg}
}

@inproceedings{
rao2021dynamicvit,
title={DynamicViT: Efficient Vision Transformers with Dynamic Token Sparsification},
author={Yongming Rao and Wenliang Zhao and Benlin Liu and Jiwen Lu and Jie Zhou and Cho-Jui Hsieh},
booktitle={Advances in Neural Information Processing Systems},
editor={A. Beygelzimer and Y. Dauphin and P. Liang and J. Wortman Vaughan},
year={2021},
url={https://openreview.net/forum?id=kR95DuwwXHZ}
}

@inproceedings{bolya2023tome,
  title     = {Token Merging: Your {ViT} But Faster},
  author    = {Bolya, Daniel and Fu, Cheng-Yang and Dai, Xiaoliang and Zhang, Peizhao and Feichtenhofer, Christoph and Hoffman, Judy},
  booktitle = {International Conference on Learning Representations (ICLR)},
  year      = {2023}
}

@INPROCEEDINGS{harari2025foveation,
  author={Gizdov, Andrey and Ullman, Shimon and Harari, Daniel},
  booktitle={2025 IEEE/CVF Conference on Computer Vision and Pattern Recognition (CVPR)}, 
  title={Seeing more with less: human-like representations in vision models}, 
  year={2025},
  volume={},
  number={},
  pages={4408-4417},
  doi={10.1109/CVPR52734.2025.00416}}

@article{kolmogorov1965,
author = {A. N. Kolmogorov},
title = {Three approaches to the quantitative definition of information * },
journal = {International Journal of Computer Mathematics},
volume = {2},
number = {1-4},
pages = {157--168},
year = {1968},
publisher = {Taylor \& Francis},
doi = {10.1080/00207166808803030},
URL = { https://doi.org/10.1080/00207166808803030},
eprint = { https://doi.org/10.1080/00207166808803030 }

}

@book{li2008kolmogorov,
  author    = {Li, Ming and Vit{\'a}nyi, Paul},
  title     = {An Introduction to {K}olmogorov Complexity and Its Applications},
  edition   = {3rd},
  series    = {Texts in Computer Science},
  publisher = {Springer},
  address   = {New York, NY},
  year      = {2008},
  isbn      = {978-0-387-49820-1},
  doi       = {10.1007/978-0-387-49820-1}
}

@inproceedings{
bardes2022vicreg,
title={{VICR}eg: Variance-Invariance-Covariance Regularization for Self-Supervised Learning},
author={Adrien Bardes and Jean Ponce and Yann LeCun},
booktitle={International Conference on Learning Representations},
year={2022},
url={https://openreview.net/forum?id=xm6YD62D1Ub}
}
